\documentclass[sigconf, nonacm=true]{acmart}

\usepackage{comment}
\usepackage{enumitem}
\usepackage{subfig}

\usepackage[ruled,linesnumbered]{algorithm2e}

\SetCommentSty{mycommfont}
\usepackage{etoolbox}
\makeatletter
\renewcommand{\SetKwInOut}[2]{%
  \sbox\algocf@inoutbox{\KwSty{#2}\algocf@typo\textbf{:}}%
  \expandafter\ifx\csname InOutSizeDefined\endcsname\relax
    \newcommand\InOutSizeDefined{}\setlength{\inoutsize}{\wd\algocf@inoutbox}%
    \sbox\algocf@inoutbox{\parbox[t]{\inoutsize}{\KwSty{#2}\algocf@typo\textbf{:}\hfill}~}\setlength{\inoutindent}{\wd\algocf@inoutbox}%
  \else
    \ifdim\wd\algocf@inoutbox>\inoutsize%
    \setlength{\inoutsize}{\wd\algocf@inoutbox}%
    \sbox\algocf@inoutbox{\parbox[t]{\inoutsize}{\KwSty{#2}\algocf@typo\textbf{:}\hfill}~}\setlength{\inoutindent}{\wd\algocf@inoutbox}%
    \fi%
  \fi
  \algocf@newcommand{#1}[1]{%
    \ifthenelse{\boolean{algocf@inoutnumbered}}{\relax}{\everypar={\relax}}%
    {\let\\\algocf@newinout\hangindent=\inoutindent\hangafter=1\parbox[t]{\inoutsize}{\KwSty{#2}\algocf@typo\textbf{:}\hfill}~##1\par}%
    \algocf@linesnumbered
  }}%
\makeatother

\usepackage{booktabs}
\usepackage{cellspace}
\usepackage{multirow}
\usepackage{multicol}
\newcommand{\tabincell}[2]{\begin{tabular}{@{}#1@{}}\linespread{0.2}#2\end{tabular}}

\newcommand{\ie}{i.e.}
\newcommand{\eg}{e.g.}

\newcommand{\algname}{CoDA\xspace}

\AtBeginDocument{%
  \providecommand\BibTeX{{%
    \normalfont B\kern-0.5em{\scshape i\kern-0.25em b}\kern-0.8em\TeX}}}

\setcopyright{acmcopyright}
\copyrightyear{2022}
\acmYear{2022}
\acmDOI{10.1145/1122445.1122456}

\begin{document}

\title{On-Device Learning with Cloud-Coordinated Data Augmentation for Extreme Model Personalization in Recommender Systems}

\author{Renjie Gu, Chaoyue Niu, Yikai Yan, Fan Wu}
\affiliation{%
  \city{Shanghai Jiao Tong University}
  \country{China}
}

\author{Shaojie Tang}
\affiliation{
   \city{University of Texas at Dallas}
   \country{USA}
}

\author{Rongfeng Jia, Chengfei Lyu}
\affiliation{%
  \city{Alibaba Group}
  \country{China}
}

\author{Guihai Chen}
\affiliation{%
  \city{Shanghai Jiao Tong University}
  \country{China}
}

\renewcommand{\shortauthors}{Gu et al.}

\begin{abstract}
Data heterogeneity is an intrinsic property of recommender systems, making models trained over the global data on the cloud, which is the mainstream in industry, non-optimal to each individual user's local data distribution. To deal with data heterogeneity, model personalization with on-device learning is a potential solution. However, on-device training using a user's small size of local samples will incur severe overfitting and undermine the model's generalization ability. In this work, we propose a new device-cloud collaborative learning framework, called \algname, to break the dilemmas of purely cloud-based learning and on-device learning. The key principle of \algname is to retrieve similar samples from the cloud's global pool to augment each user's local dataset to train the recommendation model. Specifically, after a coarse-grained sample matching on the cloud, a personalized sample classifier is further trained on each device for a fine-grained sample filtering, which can learn the boundary between the local data distribution and the outside data distribution. We also build an end-to-end pipeline to support the flows of data, model, computation, and control between the cloud and each device. We have deployed \algname in a recommendation scenario of Mobile Taobao. Online A/B testing results show the remarkable performance improvement of \algname over both cloud-based learning without model personalization and on-device training without data augmentation. Overhead testing on a real device demonstrates the computation, storage, and communication efficiency of the on-device tasks in \algname.
\end{abstract}

\ccsdesc[500]{Information systems~Information systems applications}
\ccsdesc[500]{Computing methodologies~Distributed artificial intelligence}
\ccsdesc[500]{Human-centered computing~Ubiquitous and mobile computing}

\keywords{recommender systems, data heterogeneity, model personalization, device-cloud collaborative learning, data augmentation}

\maketitle

\section{Introduction} \label{sec:introduction}

\subsection{Recommendation Model Personalization}

Recommender systems are fundamental infrastructures underlying most companies that deal with information overload and also contribute the majority of income. For example, Amazon has highlighted its personalized recommendations as its pioneered service for many years~\cite{Amazon_2020_results}; and according to Alibaba's 2021 annul report~\cite{Alibaba_2021_results}, 24\% of the year-over-year growth in customer management revenue primarily comes from recommendation feeds.

To recommend a large scale of items (\eg, billions of goods in Taobao, which is owned by Alibaba and is the largest e-commerce platform in China) for each user, recommender systems normally takes a two-stage design to balance accuracy and latency: (1) {\bf a matching stage} retrieves some (\eg, tens of thousands of) candidate items in a coarse-grained way, and (2) {\bf a ranking stage} ranks the matched items in a fine-grained way and generates the final (\eg, tens of) recommendations. The matching stage mainly adopts collaborative filtering and embedding, and further converts into a nearest neighbor search problem in the vector space. The ranking stage tends to optimize the metric of the click-through rate (CTR), while the CTR prediction model is mainly built up with embedding, multilayer perceptron (MLP), and some sequence representation structures, such as attention and gated recurrent unit (GRU). 

Typical data fields fed into the recommendation models include both user-side information (\eg, user profile and user behaviors) and item-side information (\eg, properties and categories). Meanwhile, the user-side information of different users normally differ from each other, mainly due to the distinctions in behavior patterns. In other words, users' data are non-independent and identically distributed (non-iid). For example, a user may regard shopping as a mission and perform a quick ``need-search-purchase'' process, while another user may consider shopping an enjoyment, keep browsing for a long time, and fill up the shopping cart. Such data heterogeneity causes significant challenges to the mainstream cloud-based learning method, which uses a model trained over the global data to serve each user. In particular, (1) from the aspect of model design, although the global model can represent each user with a unique user embedding vector to achieve personalized recommendation, it is prohibitive to embed personalized behavior sequences for each user, considering the large scalability of users, items, and item permutations. This brings difficulty for the model to learn different user behavior patterns and understand inner each individual user's personalized attitude towards items; (2) from the aspect of model optimization, given that the global data are a mixture of all users' data, even though a model is optimal to the global data distribution, the model cannot be optimal to any individual user's local data distribution; and (3) from the aspect of model usage, the global model is trained over all users’ data, but does inference over each user’s local data. Due to data heterogeneity, there exists discrepancy between training and test data distributions, which makes the global model not optimal for each user. These reasons motivate us to study model personalization.

To verify that the single model trained over the global data can still be significantly improved for some users' data, we trained a customized model for a cluster of users with high CTRs using only their data. The customized models took the same model structure (including user embedding) and input features as the global model. We deployed the models in the icon area of Mobile Taobao for CTR prediction (Please refer to Section~\ref{subsec:exp:scenario} for more details). Online A/B testing shows that, compared with the global model, the customized model improves CTR by 1.56\%, improves the average number of clicks per user by 3.51\%, and improves the ratio of the users who click by 1.27\%. Such remarkable improvements of online performance motivate us to study model personalization, namely, training a model for each user, which can perform accurately with regard to their local data. This new task is also intuitively called ``One Thousand Models for One Thousand Users'' in Alibaba. We note that conventional notation of personalization in recommender systems refers to different users' data as input with diverse recommendations as output, which is intuitively called ``One Thousand Faces for One Thousand Users'' in Alibaba and is supported by the single model trained over the global data on the cloud. Such data personalization with single model for personalized recommendation is parallel to the model personalization studied in this work. 

\subsection{Device-Cloud Collaborative Learning}

First, model personalization cannot be achieved with only cloud-based learning. This is because the number of models is in the same magnitude of the total number of clients (\eg, billion-scale in Taobao), and thus, maintaining such a large number of models will incur unaffordable and unacceptable overhead to the cloud servers.

Second, with the rapid proliferation and development of mobile devices, as well as the advancement of model compression techniques in the past 10 years, more and more machine learning tasks are offloaded from the cloud to the clients~\cite{jour:csur21:survey:gu}. For example, in Mobile Taobao, the deep ranking model is trained on the cloud and then deployed on mobile devices, not only to provide real-time inferences for local users, but also to avoid server congestion~\cite{10.1145/3340531.3412700}. The success of on-device inference further inspires us to consider on-device training for model personalization. However, from the perspective of each individual user, on-device training over only a small size of local data will inevitably fall into the classical dilemma of few-shot learning, generating a high generalization error and failing to realize model personalization.

Considering the infeasibility of purely cloud-based learning and on-device learning, we turn to device-cloud collaborative learning for model personalization such that the advantages of the cloud and the devices can be integrated to circumvent their disadvantages. In particular, the cloud stores the global data, the size of which is very large, to help mitigate on-device few-shot learning. Meanwhile, the ubiquitous mobile devices are quite scalable and powerful to maintain a large scale of personalized models locally and to update and use them in real time, rather than centralize the unaffordable load on the cloud. Following the principle of device-cloud collaboration, we propose a new framework where each device trains its personalized model over both the local and the outside data. The outside data are retrieved from the global data on the cloud\footnote{In practice, for privacy and security concerns, the cloud server delivers the samples after standard operations of desensitization and anonymization rather than directly sending sensitive raw data. The samples are also securely managed by the app on each mobile device and will be deleted in time after using them.} and should be similar to the local data distribution. Through cloud-coordinated data augmentation, the size of the training data on each device is enlarged, significantly improving the personalized model's generalization ability. We call our new framework ``\algname'', which is short for device-cloud \underline{Co}llaborative learning with \underline{D}ata \underline{A}ugmentation for model personalization. In nature, CoDA differs from conventional data augmentation (\eg, generative adversarial network (GAN)~\cite{GAN}) and device-cloud collaborative learning (\eg, federated learning~\cite{mcmahan2016communication}) in the existence of global data on the cloud. Specifically, conventional data augmentation considers the lack of data and generates similar fake data. In contrast, in the device-cloud collaborative scenario, although the size of the local data on each device is small, the global data on the cloud are rich enough so that there is no need to generate fake data. Regarding federated learning, its major goal is privacy preservation, and it assumes that the local user data never leave the devices, which further implies that the global data are unavailable on the cloud. The absence of global data deviates from the reality in industrial recommender systems. From this perspective, \algname is more practical. 

\begin{figure*}[!t]
    \centering
    \includegraphics[width=1.95\columnwidth]{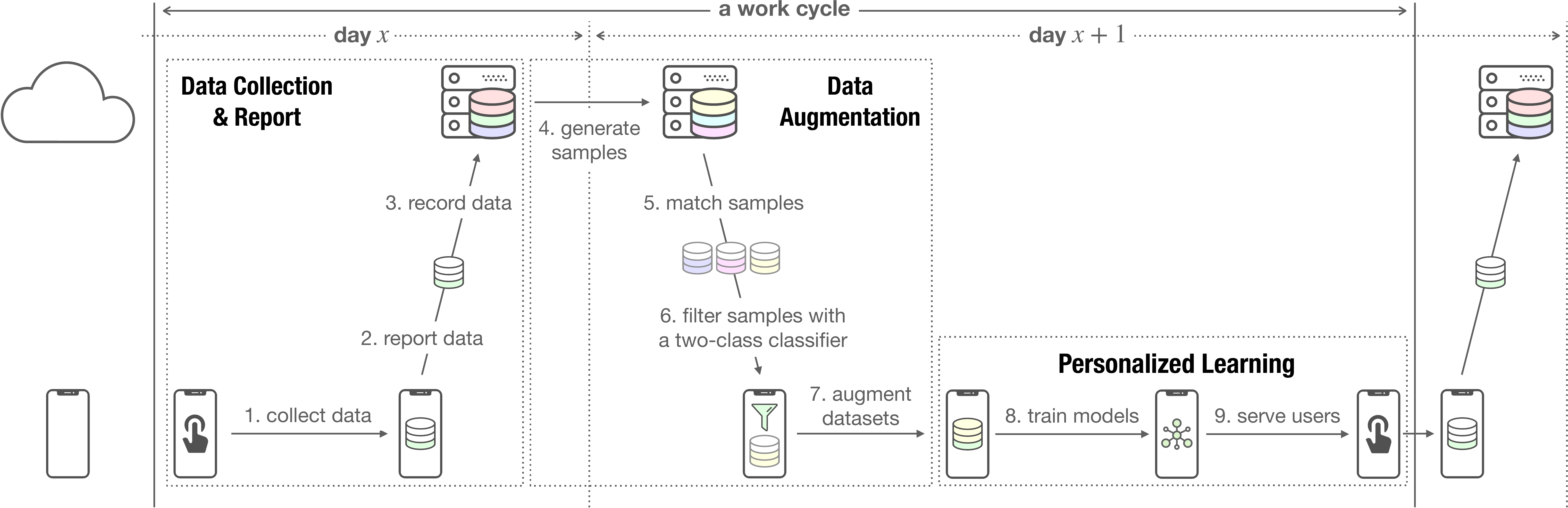}
    \caption{The workflow of \algname.}
    \label{fig:global_workflow_timeline_stage}
\end{figure*}

\subsection{Design Overview and Key Contributions}

We first give a brief overview of the algorithm design of \algname. The key regarding data augmentation for a user is how to retrieve the outside samples on the cloud that are similar to the user's local data distribution. Considering the ultra scale of the global samples on the cloud, \algname performs data augmentation in two stages, which are analogous to the matching and the ranking stages in item recommendation. The first stage is coarse-grained {\bf sample matching} on the cloud. The matching adopts the nearest neighbor search over the user feature vectors (\eg, user behavior sequence), finds a small number of similar users, and takes out their samples as the input for the second stage, the fine-grained {\bf sample filtering} on the device. In particular, each user first trains a two-class classifier to distinguish the local samples and the outside samples. After the sample classifier has converged, it will be used to score some outside samples, and those samples with scores exceeding a certain threshold will be used to augment the local samples. Based on the augmented dataset, each user will train a personalized recommendation model for local real-time serving. We note that part of the matched samples from the cloud are used for the training of the sample classifier, and another part are used for inference/scoring and filtering out the final samples for augmentation. The motivation of letting each user hold a personalized sample classifier on the device is to learn the boundary between their local data distribution and the outside data distribution automatically and accurately. 

We also build an end-to-end cloud-tunnel-device pipeline to support the flows of data, model, computation, and control in \algname. On the side of the cloud, we implement a sample preparation module for the sample matching task with the vector similarity search engine. This module also provides a sample query service based on a key-value storage system for each device to query and download the matched samples from the cloud. For the tunnel connecting the cloud and the device, it contains a down tunnel, delivering cloud-side tasks and resources to the device, and an up tunnel, reporting device-side user data to the cloud. On the side of the device, we implement the task execution module and the resource management module. The task execution module is used to execute the training and inference task scripts and is supported by a lite Python virtual machine (VM) and a mobile deep learning engine. The resources management module is implemented with the database and the file system to automatically manage the lifecycle of samples and control the version of models.

We finally overview the workflow of \algname. As shown in Figure~\ref{fig:global_workflow_timeline_stage}, each device first uploads previously collected user data to the cloud. Then, the cloud preprocesses the raw user data to generate samples and then matches samples for each target user. Each device downloads the matched samples, fine-filters them with a personalized sample classifer, and uses the filtered samples to augment the local dataset. With the augmented local dataset, each device trains a personalized recommendation model. Finally, the personalized model is used to serve the user, which also generates user data that can be used in the next work cycle.

We summarize our key contributions as follows:
\begin{itemize}
    \item To the best of our knowledge, we are the first to study model personalization in an industrial recommender system in practice, from algorithm design (Section~\ref{sec:algorithm}), system design (Section~\ref{sec:system_design}), to deployment with online performance improvement and on-device efficiency~(Section~\ref{sec:experiments}).
    \item We propose a novel device-cloud collaborative learning framework, particularly on-device personalized learning with cloud-coordinated data augmentation (\algname for short). CoDA breaks both purely cloud-based learning and on-device learning dilemmas: (1) the model trained over the global data on the cloud is not optimal to each user's local data distribution; and (2) on-device training over only the local data encounters the overfitting problem of few-shot learning and incurs a high generalization error. The key of \algname is to let each user maintain a personalized classifier on the device for data augmentation, which learns the boundary between the local data distribution and the outside data distribution.
    \item We build an online end-to-end pipeline to support the efficient flows of data, model, computation, and control between the cloud and each device.
    \item We deploy \algname in a recommendation scenario of Mobile Taobao. The online A/B testing for 23 days in two different phases shows that (1) compared with the cloud-based learning, which is the mainstream in industry, \algname improves CTR and click count by more than 1.1\%; and (2) compared with the on-device learning over only local data, \algname improves CTR and click count by more than 1.3\%. These results demonstrate the necessity of both on-device model personalization and data augmentation in \algname.
    \item We test the overhead of the on-device tasks in \algname using an iPhone 8 Plus. The evaluation results show that (1) one-round execution takes roughly 0.7s; (2) 35\% of one CPU core is used at maximum; (3) 120 MB memory is occupied; and (4) the size of downloaded data is 19.4 KB. The light load on the device reveals the practical efficiency of \algname.
\end{itemize}

\section{Related Work} \label{sec:related_work}
In this section, we briefly review the related work on recommender systems, federated learning, and other device-cloud collaborative learning frameworks, as well as data augmentation. We also clarify the key differences of our work from existing work.

{\bf Recommender Systems.} In the early development stage, the primary goal of recommender systems was to search for items or contents that the user may be interested in. The methods at this stage are filtering-based algorithms and can be generally divided into four categories~\cite{10.1016/j.knosys.2013.03.012}: content-based, demographic, collaborative, and hybrid. Content-based filtering intends to search for similar items based on what the user has viewed or rated. For demographic filtering and collaborative filtering (CF), they both assume that similar users share similar preferences for items. The former matches similar users based on their profiles (\eg, age, gender, country, etc.), and the latter matches similar users based on their historical rating behaviors. Hybrid filtering uses a combination of above methods.

When deep learning becomes popular in recent years, complementary to filtering-based methods, graph embedding-based matching methods have been proposed and gained impressive improvement in performance. \citet{proc:kdd18:EGES} focused on the billion-scale recommendation scenario in Taobao. They proposed to build an item graph from user behaviors, generate item sequences with random work, and learn item embedding with skip-gram. Based on item embedding, the pairwise similarity between any two items can be computed for matching. Later, the follow-up work \cite{proc:pakdd20:RNE} considered how to preserve the structure of the item graph and added the difference between the item distances in the original graph space and the embedding space into the loss to supervise item embedding.

As the need on recommendations becomes more and more fine-grained, a ranking stage is added after the matching stage to rank all matched items according to how much a user is interested in each candidate item, typically using CTR. Since deep learning was introduced into the ranking stage, studies have been done on designing different neural network structures for effectively predicting CTR. In the pioneering work, the Wide \& Deep structure \cite{10.1145/2988450.2988454} was proposed to combine the memorization strength of logistic regression and the generalization ability of deep neural networks (DNN). Later, Deep \& Cross network~\cite{10.1145/3124749.3124754} integrated cross network with DNN. DeepFM~\cite{10.5555/3172077.3172127} introduced factorization machine (FM) into DNN. In Alibaba, DIN~\cite{DIN} was proposed to use the attention mechanism to activate the user's historical behaviors, namely, the relative interests, with respect to the target item. In particular, DIN has gained remarkable improvement in practical recommendation performance and has been widely deployed in Alibaba's recommendation scenarios. The following-up deep interest evolution network (DIEN)~\cite{proc:aaai19:DIEN} further extracts latent interests and monitors interest evolution through GRU coupled with attention update gate. Besides the model structure, real-time response is also important in recommender systems. EdgeRec~\cite{10.1145/3340531.3412700} was proposed to offload the inference phase of the ranking model, which is trained over the global data on the cloud, to mobile devices, thus receiving user perception and system feedback in time.

Different from the single recommendation model over different users' input data with personalized outputs in existing work, which is also the mainstream in industry, we consider model personalization in recommender systems, which requires on-device training but meets the overfitting problem. The proposed \algname mitigates overfitting through cloud-coordinated data augmentation.

\begin{figure*}[!t]
    \centering
    \includegraphics[width=1.95\columnwidth]{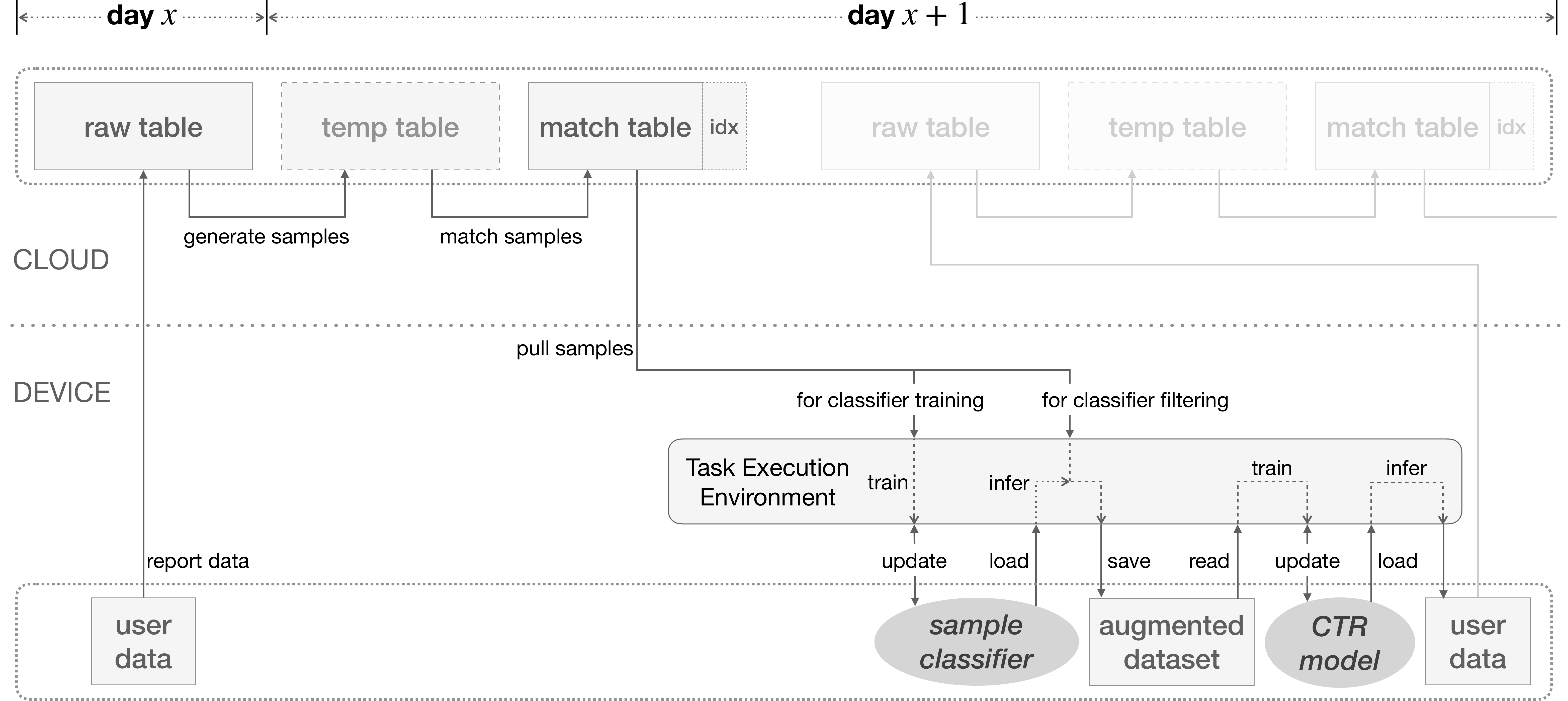}
    \caption{The dataflow of \algname.}
    \label{fig:dataflow}
\end{figure*}

{\bf Federated Learning over Non-iid Data.} Federated learning \cite{mcmahan2016communication} and the underlying federate averaging algorithm (FedAvg) were first proposed by Google, which aim at coordinating a large scale of mobile devices to train a global model without transferring local user data to a centralized server, so that data privacy can be well protected. In particular, each round of FedAvg restarts a machine learning process and averages the model updates of participating devices to update the global model. Under the federated learning framework, much work studies how to guarantee model convergence and improve model performance over non-iid data. \citet{pmlr-v97-yu19d} (resp., \citet{proc:iclr19:zhihuazhang:convergence}) analyzed the convergence of FedAvg when all the clients are always available and fully (resp., partially) participate in the collaborative training. \citet{proc:icml20:scaffold} proposed a variant of FedAvg, called stochastic controlled averaging, which incorporates the gradients in the previous round to adjust the global direction of federated optimization and guarantees good convergence. \citet{zhao2018federated} analyzed the negative effect of data heterogeneity on model performance and proposed to share a small subset of global data among mobile devices. \citet{pmlr-v97-eichner19a} captured data cyclicity underlying federated learning and offered a pluralistic multi-model solution for convex objectives and sequential stochastic gradient descent (SGD). \citet{pmlr-v97-mohri19a} considered an unfairness issue that the global model can be unevenly biased toward different users. They thus proposed a new agnostic federated learning framework where the global model can be optimized for any possible target distribution, which is formed via a mixture of client distributions.

In addition to the federated optimization of a global model, much recent work has turned to federated learning for model personalization. \citet{proc:nips17:smith:fed:multitask} focused on learning separate but related personalized models for different clients by leveraging multi-task learning for shared representation. \citet{chen2018federated} instead adopted meta-learning to enable client-specific modeling, where clients contribute information at the algorithm level rather than the model level to help train the meta-learner. Several following work \cite{jiang2019improving, fallah2020personalized, NEURIPS2020_24389bfe, 9355664} also proposed to combine federated learning with meta-learning for personalization. \citet{deng2020adaptive} demonstrated that federated learning has both an innate advantage and an urgent need of personalization due to data heterogeneity. They proposed to leverage an adaptively updated parameter to tune the personalization degree for each user. \citet{mansour2020three} recently made an in-depth study on the necessity and feasibility of personalization in federated learning. They proposed three novel and general frameworks for personalized federated learning, namely, user clustering, data interpolation, and model interpolation.

The major difference between federated learning and \algname is the setting on the global data distribution. Federated learning assumes that user data are kept only on user devices, and no data are available on the cloud. The cloud and each device exchange model parameters in federated learning. However, in many real-world industrial scenarios, most user data have already been uploaded to the cloud. In this practical setting, we re-examine the roles of the cloud and the device, and propose on-device learning with cloud-coordinated data augmentation for model personalization.  

{\bf Collaborative Learning.} Device-cloud collaborative learning intends to integrate the resources of the mobile devices and the cloud to accomplish a learning task. Besides the popular federated learning framework, which offloads the whole training task from the cloud to mobile devices, some other collaborative learning work studied how to allocate tasks in a different and more reasonable way. JointDNN~\cite{8871124} realized collaborative learning by partitioning the model into two parts and making the cloud and the mobile device each be responsible for the computation of one part. COLLA~\cite{10.1145/3318216.3363304} used a bidirectional knowledge distillation to transfer information between the cloud model and the device model. D-Distillation~\cite{NEURIPS2020_DistributedDistillation} proposed to use a public reference dataset to share model outputs among neighboring devices for distillation, while each device's own dataset is kept private. DCCL~\cite{10.1145/3447548.3467097} designed a device-cloud collaborative learning framework for recommender systems, which leverages model patch learning~\cite{proc:sigir20:patch} for device-side model penalization and knowledge distillation for cloud-side model update. 

Parallel to the model-based collaborative learning work above, \algname is sample-based and relies on cloud-coordinated data augmentation to facilitate on-device personalization learning.

{\bf Data Augmentation.} Data augmentation has been widely used in the area of computer vision to deal with the problems of few-shot learning. Traditional data augmentation methods use basic image manipulations (\eg, cropping, flipping, and rotation) to generate new resemble samples by operating over the original ones. Advanced methods such as kernel filter~\cite{kang2017patchshuffle}, noise injection~\cite{moreno2018forward}, and random erasing~\cite{zhong2020random} share a similar idea with the traditional methods, but adopt more complicated transformations. As a combination of these ideas, AutoAugment~\cite{cubuk2018autoaugment} leveraged neural architecture search (NAS) to automatically search for a proper composition of image manipulations which can apply to the target dataset well. Different from the above methods, GAN~\cite{GAN} was proposed to learn the data distribution with a min-max game, where a generative network generates fake samples whereas the discriminative network intends to distinguish the fake samples from real samples. Later, GANs with different structures were proposed, such as supervised GAN (CGAN)~\cite{tc:arxiv14:CGAN}, CNN + GAN (DCGAN)~\cite{proc:iclr16:DCGAN}, Wasserstein distance + GAN (WGAN)~\cite{proc:icml17:WGAN}, and transformer + GAN (TransGAN)~\cite{tc:arxiv21:TransGAN}. As GAN had proved its effectiveness for data augmentation, it was also introduced to recommender systems to deal with the dataset problems met by CF. ${\rm {RAGAN}^{BT}}$~\cite{proc:www19:RAGANBT} was proposed to generate plausible ratings with GAN for handling the data sparsity problem in CF. AugCF~\cite{10.1145/3292500.3330873} also focused on the data sparsity problem and built a GAN-based end-to-end CF model, which uses its generative network to augment the dataset in the first stage and uses its discriminative network to perform CF in the second stage. AR-CF~\cite{proc:sigir20:AR-CF:gan} considered the cold-start problem in CF and used GAN to generate fake neighbors for new users and items.

Different from existing data augmentation work which deals with the small dataset problem normally through generating fake data, \algname is under a practical setting where the global dataset on the cloud is rich and sufficient. Thus, we propose to relieve the lack of each user's local dataset by augmenting with the real samples retrieved from the global pool.
\section{Algorithm Design} \label{sec:algorithm}
In this section, we first introduce the overall dataflow in \algname to show how user data are used to finally accomplish model personalization. We then present the design of the core algorithms in sequence, including cloud-based sample matching, on-device sample filtering, and on-device training over augmented data.

\subsection{Dataflow}

As shown in Figure~\ref{fig:dataflow}, user data are first generated and collected on mobile devices during each user's interaction with the app. Then, the data are reported to the cloud and stored in the raw data table. The sample generation step uses the raw data table as input, preprocesses the data, and outputs samples to a temporary sample table. This step not only desensitizes the raw user data, but also provides necessary inputs (\ie, samples and user features) for the following step of sample matching. After that, the cloud retrieves samples for each user, saves the results to the matched sample table, and additionally build an index table for efficient sample pulling.

In the next day, each device first pulls the matched samples from the cloud and divides them into two subsets: one is for the training of a personalized sample classifier, and the other is for classifier filtering. After training the classifier with a subset of the matched samples, the rest samples are scored and filtered by the classifier and saved to the local storage to augment the local dataset. When the size of the augmented dataset reaches a preset threshold, the samples are read out and used to train the recommendation model (\eg, a CTR prediction model for ranking). The updated model is then used to serve the user in future requests, during which new user data are generated.

\begin{algorithm}[!t]
\caption{KNN-Based Sample Matching}
\label{alg:sample_matching}
\SetAlgoLined
\SetNoFillComment
\DontPrintSemicolon
\SetKwInOut{KwData}{Data}\SetKwInOut{KwInput}{Input}\SetKwInOut{KwOutput}{Output}
\SetKwProg{Initialization}{initialization:}{}{}
\SetKwProg{Fn}{Function}{:}{\textbf{End Function}}
\SetKwFunction{FextractFeature}{extractFeature}

\KwData{global user set $U$, global sample set $S$, the local sample set $S_u \subset S$ of user $u \in U$}
\KwInput{target users $U_t \subset U$, parameter $K$ in KNN}
\KwOutput{matched sample set $\dot{S}_{\alpha}$ for each target user $\alpha \in U_t$}
\BlankLine

\Fn{\FextractFeature{{\rm sample} $s$}}{
    a feature vector $v \in s$ to represent the user, \eg, the user profile vector or the user behavior sequence in $s$\;
    \KwRet $v$ 
}
\BlankLine

\tcc{build the map between user ID and user vector as well as the global user vector space}
initialize the user ID to user vector map \texttt{user\_vector\_map}\;
initialize the global user vector space $V \leftarrow \emptyset$\;
\ForEach{$u \in U$}{
    user vector $v_u \leftarrow \frac{1}{|S_u|} \sum\limits_{s \in S_u}\mathrm{\texttt{extractFeature}}(s)$\;
    ${\rm user\_vector\_map[u]} \leftarrow v_u$\;
    $V \leftarrow V \cup \{v_u\}$\;
}
\BlankLine

\tcc{do sample matching for target users}
\ForEach{$\alpha \in U_t$}{
    $v_\alpha \leftarrow {\rm \texttt{user\_vector\_map}[\alpha]}$\;
    matched user set $\dot{U}_{\alpha} \leftarrow$ \{$u$ | $v_u$ is one of the $K$ nearest neighbors of $v_{\alpha}$ in $V$\}\;
    matched sample set $\dot{S}_{\alpha} \leftarrow \bigcup\limits_{u \in \dot{U}_{\alpha}} S_u$\;
}
\end{algorithm}

\subsection{Cloud-Based Sample Matching} \label{subsubsec:alg_sample_matching}

The sample matching algorithm, as the cloud-side part of data augmentation, uses the global set of samples collected from all users as the data source, and retrieves some candidate similar samples for each user. As shown in Algorithm~\ref{alg:sample_matching}, the cloud first performs feature extraction to obtain user vectors. The feature extraction mainly takes a user's sample as its input, extracts a user-related feature (\eg, the user profile or the user behavior sequence) from the sample, and outputs a vector for representing the user. Then, for each target user, the cloud searches for $K$ nearest neighboring (KNN) users in the whole user vector space and takes out these $K$ users' samples as the matched ones. We note that the sample matching algorithm uses the contents of a user's samples to represent the user and performs a user-level matching, which in fact can be regarded as matching samples in a batch mode. In addition, some other methods of user matching or behavior sequence matching can also be used as substitutable solutions.

\subsection{On-Device Sample Filtering} \label{subsubsec:alg_sample_filtering}

\begin{algorithm}[!t]
\caption{Classifier-Based Sample Filtering}
\label{alg:sample_filtering}
\SetAlgoLined
\SetNoFillComment
\DontPrintSemicolon
\SetKwInOut{KwData}{Data}\SetKwInOut{KwInput}{Input}\SetKwInOut{KwOutput}{Output}
\SetKwProg{Initialization}{initialization:}{}{}
\SetKwProg{Fn}{Function}{:}{\textbf{End Function}}
\SetKwFunction{FextractFeature}{extractFeature}

\KwData{$u$'s local sample set $S_u $}
\KwInput{$u$'s matched sample set $\dot{S}_u$, \\
        a filtering threshold $\sigma \in [0, 1]$}
\KwOutput{$u$'s filtered sample set $\ddot{S}_u$}
divide $\dot{S}_u$ into $\dot{S}_u^1$ for training and $\dot{S}_u^2$ for inference\;
\BlankLine
\tcc{train the sample classifier on the device}
\If{{\rm the sample classifier $C_u$ is not initialized}}{
randomly initialize $C_u$\;
}
train $C_u$ over $S_u \cup \dot{S}_u^1$ with the loss function $L_u$\;
\BlankLine
\tcc{use the classifier to filter samples}
initialize $\ddot{S}_u \leftarrow \emptyset$\;
\ForEach{$s \in \dot{S}_u^2$}{
    \If{$C_u(s) \ge \sigma$}{
        $\ddot{S}_u \leftarrow \ddot{S}_u \cup \{s\}$\;
    }
}
\end{algorithm}

The sample filtering algorithm, as the device-side part of data augmentation, takes some of the outside matched samples and the local samples as input, trains a personalized two-class classifier for each user, and then uses the trained sample classifier to score and filter out the rest outside samples for data augmentation.

For each user $u$, we let $S_u$ denote $u$'s local samples, let $\dot{S}_u$ denote the matched samples for $u$ from the cloud, part of which $\dot{S}_u^1$ is used for the training of the sample classifier, and the other part $\dot{S}_u^2$ is used for the inference/filtering of the sample classifier. Then, the whole training set of the sample classifier is $S_u \cup \dot{S}_u^1$, which is denoted as $\hat{S}_u$. For each training sample $(x, l) \in \hat{S}_u$, $x$ is the feature vector, and $l \in \{0, 1\}$ labels whether the sample is outside or local (\ie, 0 for outside and 1 for local). The cross-entropy loss function of the user $u$'s sample classifier $C_u$ is defined as:
\begin{align} \label{eq:classifier_loss}
L_u = -\frac{1}{\left|\hat{S}_u\right|} \sum_{(x, l) \in \hat{S}_u} [ l \log C_u(x) + (1 - l) \log (1 - C_u(x)) ].
\end{align}
The optimization objective of the sample classifier is to minimize the cross-entropy loss, which implies that the local samples should be scored with values closer to 1, while the outside samples should be scored with values closer to 0. In addition, for a trained sample classifier, the outside samples with high scores (\ie, false positives) are considered to be similar to the local samples and should be used for augmentation. Intuitively, the sample classifier functions as a discriminative model to learn the boundary between the local data distribution and the outside data distribution.

Based on the personalized sample classifier, we present the detailed sample filtering algorithm in Algorithm~\ref{alg:sample_filtering}. The algorithm is divided into two stages: classifier training and sample filtering. Each device first trains the sample classifier with both the local samples and a subset of the outside matched samples. Then, each device uses the trained classifier to filter the rest matched samples. The filtering threshold score $\sigma$ controls the decision boundary. Those outside samples, whose predicted scores are larger than $\sigma$, will be considered to share a similar distribution with the local samples, and therefore are kept on the device to augment the local dataset. At the end, each user $u$'s on-device augmented dataset $S'_u$ is denoted as $S_u \cup \ddot{S}_u$.

\subsection{On-Device Training over Augmented Data}\label{subsec:sys_model_training}

\begin{algorithm}[!t]
\caption{Personalized Model Training}
\label{alg:model_training}
\SetAlgoLined
\SetNoFillComment
\DontPrintSemicolon
\SetKwInOut{KwData}{Data}\SetKwInOut{KwInput}{Input}\SetKwInOut{KwOutput}{Output}
\SetKwProg{Initialization}{initialization:}{}{}
\SetKwProg{Fn}{Function}{:}{\textbf{End Function}}
\SetKwFunction{FextractFeature}{extractFeature}

\KwData{$u$'s local sample set $S_u$,\\ $u$'s augmented dataset $S'_u := S_u \cup \ddot{S}_u$}
\KwInput{$u$'s initialized recommendation model $M_u$}
\KwOutput{$u$'s trained recommendation model $M_u$}
divide $S_u$ into $S_u^1$ for training and $S_u^2$ for validation\;
validate $M_u$ on $S_u^2$ and get the model accuracy $ACC$\;
\ForEach{mini-batch {\rm in the training set} $S_u^1 \cup \ddot{S}_u$}{
    optimize $M_u$ on \textit{mini-batch} and get $M_{u}'$\;
    validate $M_{u}'$ on $S_u^2$ and get $ACC'$\;
    \If{$ACC' > ACC$}{
        $M_u \leftarrow M_{u}'$\;
        $ACC \leftarrow ACC'$\;
    }
}
\end{algorithm}

After data augmentation, each device can train a personalized recommendation model (\eg, a CTR prediction model) to serve the user. We present the process of on-device training over augmented data in Algorithm~\ref{alg:model_training}. First, the augmented dataset $S_u = S_u \cup \ddot{S}_u$ will be split into a training set $S_u^1 \cup \ddot{S}_u$ for optimizing the recommendation model as well as a validation set $S_u^2$ for testing the model and determining whether to keep the updated model. In particular, to ensure that the recommendation model is updated towards the optima over a user's local data distribution, only the user's local samples should be used for model validation (\ie, $S_u^2 \subset S_u$), which can mitigate the possible bias in the filtered outside samples. Additionally, after each iteration of local training over a batch of training samples, if the model accuracy (\eg, the metric of area under the curve (AUC) widely used in evaluating recommendation performance) increases, then the recommendation model will be updated.

\begin{figure*}[tb]
    \centering
    \includegraphics[width=0.9\linewidth]{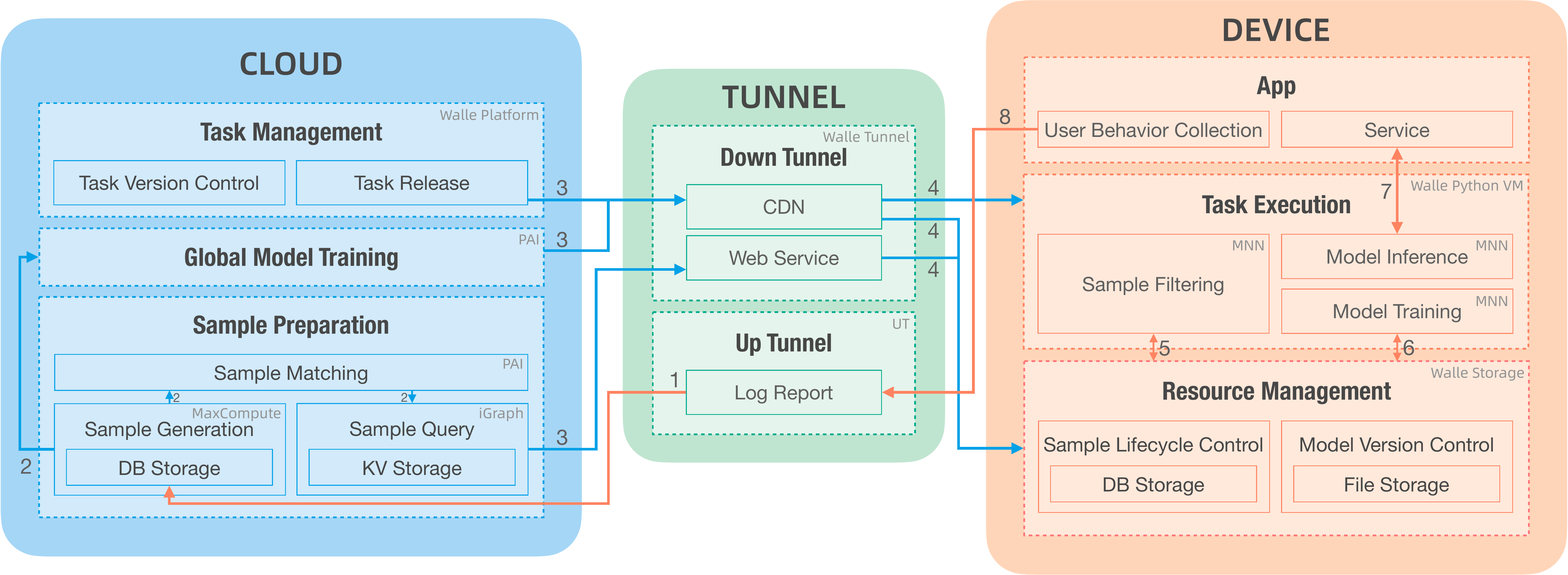}
    \caption{The system architecture of \algname.}
    \label{fig:system_modules_workflow}
\end{figure*}

\section{System Design} \label{sec:system_design}

Although the pipeline of conventional cloud-based learning is off-the-shelf, the core modules of \algname's new device-cloud collaborative learning need to be built from scratch. The key challenges include (1) for the cloud, the efficient storage, fast query, and durative distribution of matched samples at the level of user/device and the need to support a large number of devices; (2) for each device, the fine-grained management of task resources~(\ie, samples, models, and scripts) under the strict constraint of local storage, as well as the process control and the conflict control of multiple on-device tasks at the levels of sample and model.

We design the system architecture of \algname as shown in Figure~\ref{fig:system_modules_workflow}. The end-to-end pipeline is divided into 3 parts: cloud, tunnel, and device. Each part is made up of several modules, which cooperate with each other to realize the algorithm design of \algname. Some existing basic platforms and frameworks in Alibaba Cloud and Mobile Taobao, the names of which are put on the top-right corner and colored in gray, are used to support the implementation of the system modules. We list their names and functionalities in Table~\ref{tab:basic_platforms}.


\begin{table}[!t]
    \centering
    \caption{Basic platforms and frameworks in Alibaba Cloud and Mobile Taobao underlying \algname.}
    \label{tab:basic_platforms}
    \resizebox{\linewidth}{!}{
    \begin{tabular}[t]{llp{0.7\linewidth}}
        \toprule
         & \multicolumn{1}{c}{\textbf{Name}} & \multicolumn{1}{c}{\textbf{Functionality}}\\
        \midrule
        & MaxCompute~\cite{MaxCompute} & A data warehousing and processing platform.\\
        \cmidrule{2-3}
        \multirow{4}{*}{\textbf{Cloud}} & \multirow{2}{*}{PAI~\cite{PAI}} & A platform of artificial intelligence providing Python data science environment.\\
        \cmidrule{2-3}
        & iGraph & An online graph storage and query system.\\
        \cmidrule{2-3}
        & Walle Platform & A platform for managing on-device tasks.\\
        \midrule
        \multirow{3}{*}{\textbf{Tunnel}} & \multirow{2}{*}{Walle Tunnel} & A down tunnel for sending device-computing task resources from the cloud to devices.\\
        \cmidrule{2-3}
        & UT & An up tunnel for reporting device log.\\
        \midrule
        & \multirow{2}{*}{Walle Storage} & A framework for managing the storage of device-computing tasks. \\
        \cmidrule{2-3}
        \multirow{2}{*}{\textbf{Device}} & \multirow{2}{*}{Walle Python VM} & A Python environment to execute device-computing tasks. \\
        \cmidrule{2-3}
        & \multirow{2}{*}{MNN~\cite{alibaba2020mnn}} & A mobile deep learning framework for on-device model training and inference.\\
        \bottomrule
    \end{tabular}%
    }
\end{table}%

\subsection{Work Cycle}

We first introduce the work cycle throughout all the system modules: (1) user data are reported to the sample preparation module through the up tunnel; (2) in the sample preparation module, user data are processed to samples. Samples are then matched for each target user and stored in a key-value storage to provide sample query service. Meanwhile, the global model training module can train a cloud-based model using the global samples if needed; (3) from the cloud side, the resources for the on-device learning tasks, including matched samples, the initial global model, and task-related scripts, are delivered to mobile devices through the down tunnel; (4) the resources are sent to the on-device resource management module and the task execution module. In particular, the matched outside samples are saved to the on-device database storage with their life-cycle managed, the model is saved to the file storage with its version controlled, and the task-related scripts are sent to the task execution module; (5) the on-device sample filtering is executed in the task execution module, interacts with the resource management module to train the sample classifier, and uses the trained classifier to filter out the outside samples to augment the local dataset; (6) the on-device personalized model training is also executed in the task execution module, interacts with the resource management module to optimize the recommendation model with the augmented dataset; (7) supported by the background model inference task run in the task execution module, the app provides real-time recommendation service for the user; (8) during the interaction between the user and the app, the user behavior collection module collects new user data and uploads them through the up tunnel.

In what follows, we describe the system modules in detail. We mainly introduce the key functionalities and design principles of each module, as well as the underlying platforms and frameworks.

\subsection{Cloud-Side Modules}

On the side of the cloud, sample preparation, global model training, and task management are three key modules.

\begin{figure}[!t]
    \centering
    \includegraphics[width=0.95\linewidth]{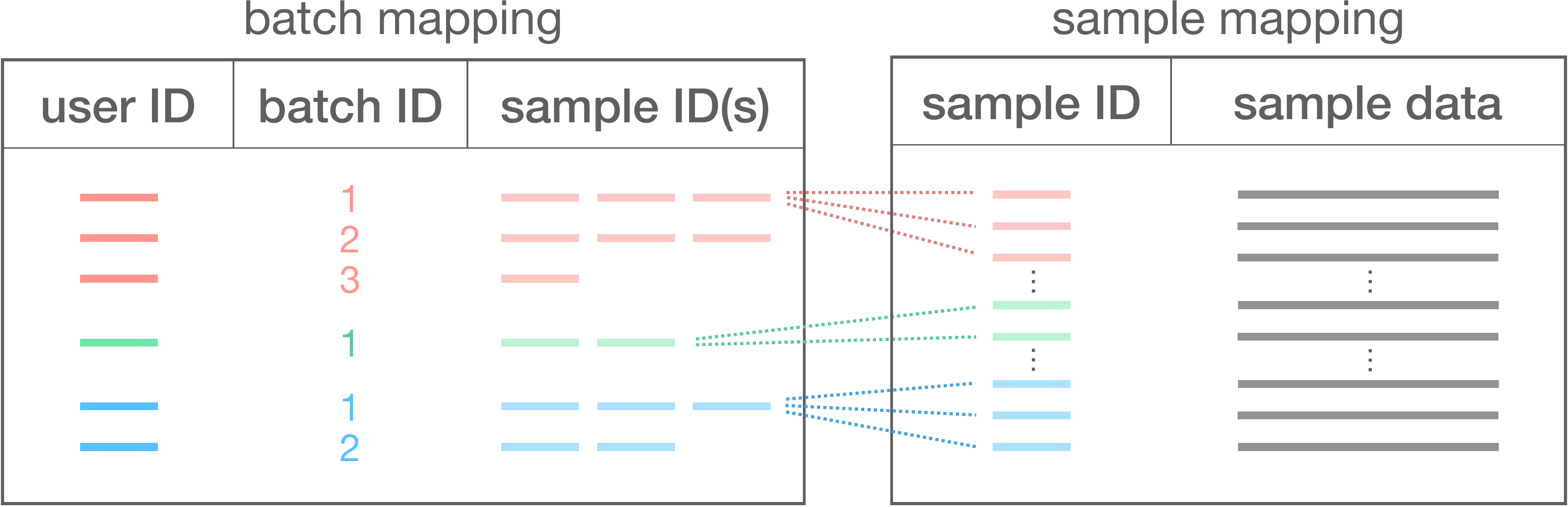}
    \caption{Two mappings used to support the sample query.}\label{fig:cloud_matching_table}
\end{figure}

\subsubsection{Sample Preparation}

This module supports the sample matching algorithm on the cloud. It is made up of three sub-modules: sample generation, sample matching, and sample query.

\textbf{Sample Generation.} The sample generation sub-module takes user data as input, performs massive data processing, and outputs generated samples. The underlying platform used to support this module is MaxCompute~\cite{MaxCompute}, which is a large-scale data warehousing and processing platform. We make use of its database-like storage and distributed computing ability to efficiently store massive amounts of structured user data and process them in parallel.

\textbf{Sample Matching.} The sample matching sub-module prepares similar samples for each user. To facilitate fast matching for huge amounts of samples, we build indexes for the matching table to support efficient sample lookup, and also use the vector similarity search engine, called Faiss from Facebook~\cite{Faiss}, to do fast similarity calculation in KNN. These 
accelerations are performed in a Python data science environment supported by Alibaba's platform of artificial intelligence (PAI) \cite{PAI}.

\textbf{Sample Query.} The sample query module allows a mobile device to query a batch of samples at one time. After the sample matching step, the matched samples for a target user are split into several batches. Each batch is assigned with a batch ID for query. This batch query mechanism avoids the issue that the entire set of the matched samples can be too large in size to be sent to the device at once. Another practical and significantly important issue is data duplication. The KNN-based sample matching step may match one sample to many different users, introducing a lot of sample duplicates in the overall matching result. If the matched sample data is directly stored for each user, the size of system data will grow to roughly $(K + 1)$ times its original size. To deal with this problem, we assign a unique ID for each sample and maintain a mapping from sample IDs to sample data. Only the matched sample IDs are saved for each user, which greatly reduce the additional space cost. The sketches of the batch mapping (from batch ID to sample IDs) and the sample mapping (from sample ID to sample data) are shown in Figure~\ref{fig:cloud_matching_table}. The two mappings are stored in a key-value form in iGraph, which is a large-scale online graph storage and query system supporting high-efficient lookup for key-value pairs.

\subsubsection{Global Model Training}
The global model training module supports training a model over the global data on the cloud, which is conventional non-personalized learning and is the current mainstream in industry. We leverage distributed TensorFlow~\cite{tensorflow2015-whitepaper} as the cloud-side machine learning framework. The large amount of computing resources (\ie, CPU, GPU, and memory) and the distributed computing ability are provided by PAI. In addition, since the on-device execution of the model is implemented using mobile neural network (MNN)~\cite{alibaba2020mnn} (which will be introduced in Section~\ref{subsubsec:sys_task_execution}), the trained TensorFlow model is converted to an MNN model with the help of MNNConvert. During the conversion, the model's structure is converted and simplified to compress its size for efficiency.

\subsubsection{Task Management}
The task management module is the entrance for machine learning engineers to submit on-device tasks. Walle platform, which is the cloud-side part of the Walle device-computing framework, works as the underlying platform and can control and monitor the overall status of the device-computing tasks. Two core sub-modules of task management are task version control and task release.

\textbf{Task Version Control.} Task version control is implemented using Git. It records the whole submission history of a task so that each engineer can easily figure out the changes between different task versions. It also allows the system to do a quick version rollback in case the new version of the task does not function as expected.

\textbf{Task Release.}
The task release sub-module functions like a router. It not only determines whether a task is sent to devices, but also controls which devices the task will be sent to. Once a device becomes online, this sub-module will generate an active task list for it according to both its device information (\eg, user ID, device ID, app version) and the hitting rules defined in the task configuration. Then, the device compares the new active task list with its local task list, removes the outdated tasks, and fetches the newly-activated ones. This dynamic task release mechanism brings flexibility to the system. Furthermore, by combining the task version rollback function with the dynamic release feature, a problematic task release can be easily revoked. As a result, the stability of the system can be better ensured.

\subsection{Tunnel Connecting Cloud and Device}

The tunnel connecting the cloud and the device is an important bridge for communication. Based on the direction of messages, the tunnel can be divided into the down tunnel and the up tunnel.

\subsubsection{Down Tunnel}
The down tunnel is used to send data from the cloud to devices. For the static resources (\eg,~task scripts and models) which remain same for all accessing devices, they are cached on a content delivery network (CDN) to accelerate the delivery as well as reduce the server load. For the dynamic resources (\eg, the matched samples) which differ from device to device, a web service is provided to response authenticated device queries. Walle Tunnel encapsulates the above-mentioned two delivery methods and works as the main delivery tunnel of the system.

\subsubsection{Up Tunnel}
During on-device task execution, user log is generated and reported to the cloud through the up tunnel. The report tunnel is supported by user track (UT). To improve system efficiency, log messages are first aggregated on the device and then reported to the server. The whole reporting process is encrypted using secure sockets layer (SSL) to guarantee data security.

\subsection{Device-Side Modules and App}

On the side of the device, resource management and task execution are two key modules. These modules also support the mobile app on the top to serve and interact with the user in real time.

\subsubsection{Resource Management} \label{subsubsec:sys_resource_management}

The word ``resource'' here mainly refers to data resources. Considering on-device storage is quite limited, unnecessary space occupation can bring negative effects on user experience. Therefore, we design and implement a strict resource management module to take care of the on-device data. The implementation is based on the basic application programming interfaces (APIs) (\eg, database SQL execution and file operations) provided by Walle Storage.

\begin{figure}[!t]
    \centering
    \includegraphics[width=1.0\linewidth]{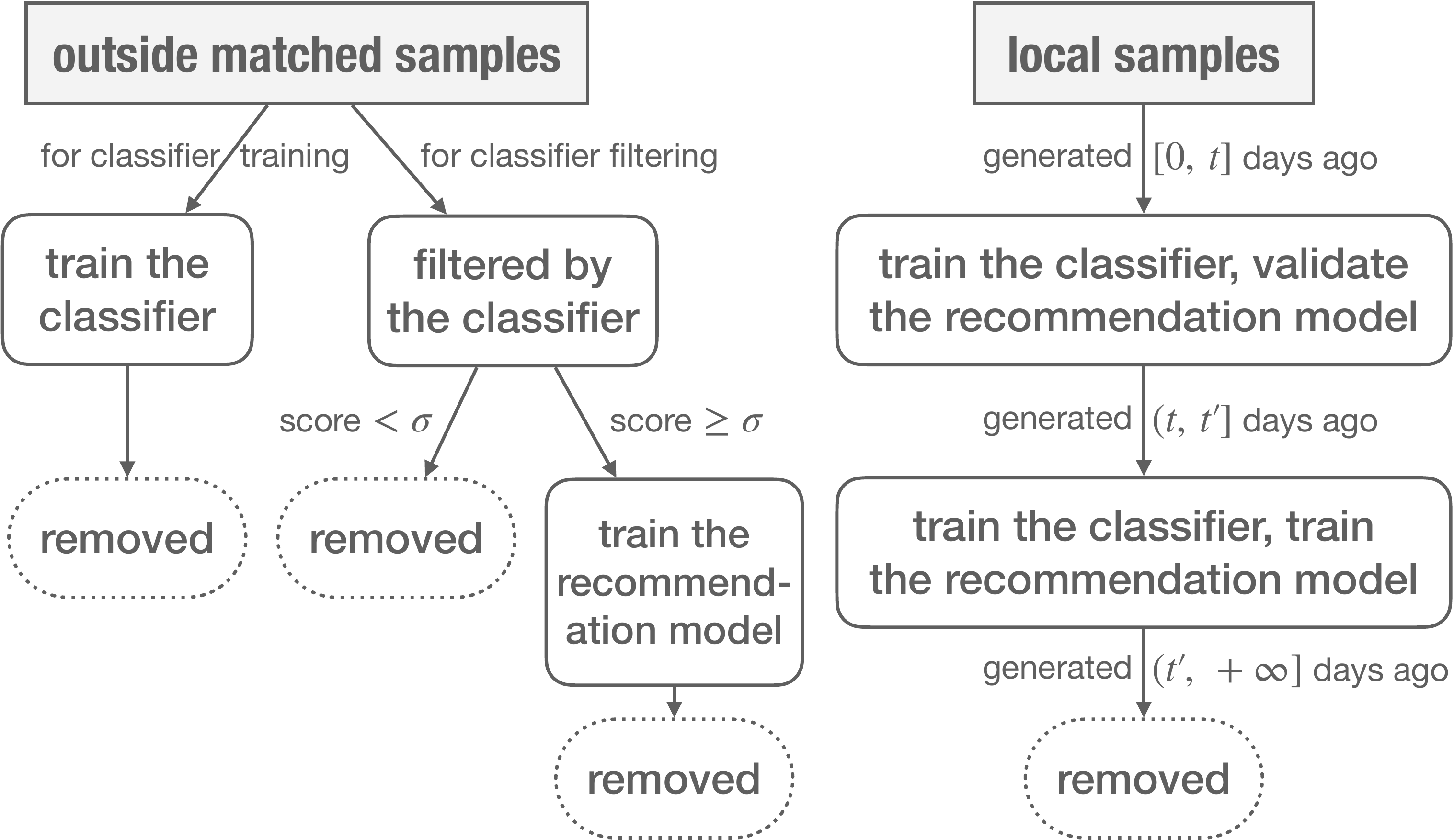}
    \caption{The policies of sample lifecycle control.}
    \label{fig:sample_lifecycle}
\end{figure}

\textbf{Sample Lifecycle Control.}
Samples are stored in an on-device database with their lifecycle managed. We take different control policies for the outside matched samples and the local samples, which are stored in two different tables. The sample lifecycle control policies are shown in Figure~\ref{fig:sample_lifecycle}.

For the outside matched samples, they can be divided into the following categories according to their usages in data augmentation. If a sample is used for the training of the sample classifier, it will be removed after training. In contrast, if a sample is used for being filtered by the sample classifier, (1) if the sample gets a score of $[0, \sigma)$, then it will be abandoned; (2) otherwise, it will be used to augment the local dataset for the training of the personalized recommendation model. After usage, the sample will be removed from the database.

For the user's local samples, they are stored in another table with a different lifecycle. If a sample was generated $[0, t]$ days ago, it will be used for the training of the sample classifier and the validation of the recommendation model. If a samples was generated $(t, t']$ days ago, it will be
used for the training of the sample classifier and the training of the recommendation model. If a sample was generated $(t', +\infty]$ days ago, it will be removed. 

In most cases, the above lifecycle control mechanism is sufficient for sample management. In case that an unexpected large amount of samples are received, we set a size limit for each of the two sample tables. If the table size reaches the limit, a forced cleanup will be triggered. The cleanup rules are (1) for the outside matched samples, the ones with lower scores predicted by the sample classifier will be removed until the table size decreases by half. We note that if a sample has not been scored by the sample classifier, it will be scored with 0 by default; and (2) for the user's local samples, the ones with earlier generation time will be removed until the table size decreases by half. The forced cleanup rules can guarantee that the total size of sample storage never exceeds the predefined limit.

\begin{figure}[!t]
    \centering
    \includegraphics[width=0.7\linewidth]{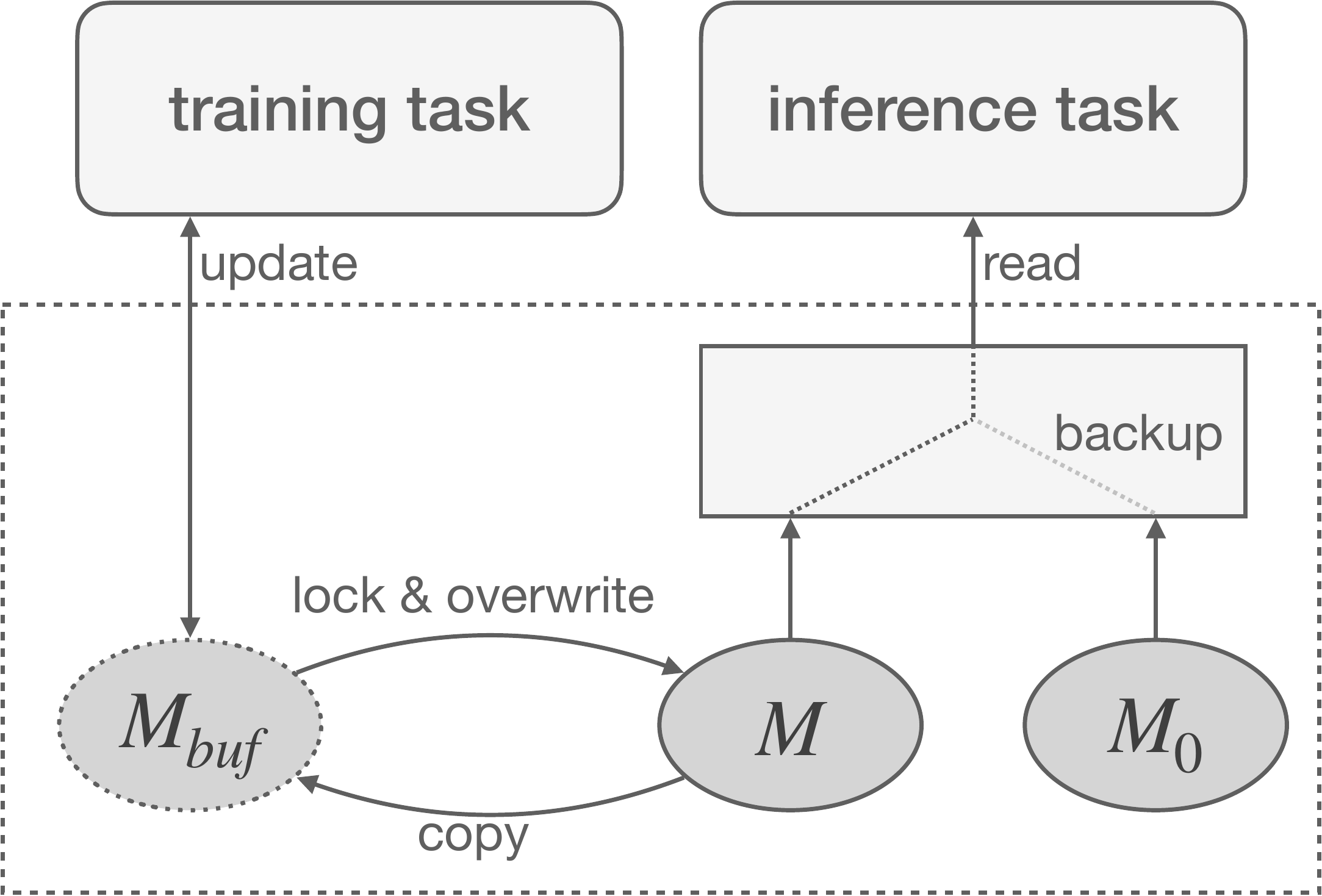}
    \caption{The mechanism of model version control.}
    \label{fig:model_version_control}
\end{figure}

\textbf{Model Version Control.} Models (\ie, the sample classifier and the recommendation model in \algname) are stored as files with their versions controlled. The design of model version control sub-module needs to meet the following requirements: (1) being storage-friendly; (2) supporting quick rollback when the accuracy of the updated model decreases; and (3) avoiding the write–read conflict for the training task and the inference task. Guided by these requirements, we design the on-device model version control mechanism as shown in Figure~\ref{fig:model_version_control}. The key steps are summarized as follows: (1) initialize/receive and save a model $M$; (2) make a duplicate for $M$ and name it as $M_0$. $M_0$ serves as an always-available backup model, which is never modified after its creation; (3) when the on-device model training starts, make a duplicate for $M$ and name it as $M_{buf}$. Provide the training task with $M_{buf}$, serving as the buffer for the training process; (4) when the model model training finishes, check whether a rollback is needed. In particular, (4.1) if the updated model needs to be saved, add a write lock for $M$ and overwrite $M$ with $M_{buf}$. Then, release the write lock and remove $M_{buf}$; (4.2) if a rollback is needed, then remove $M_{buf}$; (5) when model inference starts, provide the inference task with $M$, if there is no write lock; otherwise, provide it with $M_0$.

We note that in most of the time, only two versions of model (\ie, $M_0$ and $M$) are maintained on the device, while $M_{buf}$ just exists during the short period of on-device training. Therefore, our model version control mechanism is storage-friendly. The second requirement of quick rollback is supported by Step 4. The third requirement of avoid write–read conflict is met by Step 5. Furthermore, the whole process of model version control is transparent to the task execution, which simplifies the task design.

\subsubsection{Task Execution} \label{subsubsec:sys_task_execution}
The task execution module contains a lite Python environment provided by Walle Python VM. All the computing resources (\eg, CPU and memory) allocated to the tasks are managed by Walle Python VM automatically. In addition, MNN~\cite{alibaba2020mnn}, as a mobile deep learning framework, has been integrated into the on-device Python environment with its Python APIs exposed to the task scripts. With the help of MNN, each mobile device can perform model inference and model training to accomplish the following three on-device tasks: sample filtering, the training and the inference of personalized recommendation model. The triggers of these tasks are managed by the task execution module. Once a new batch of matched samples is received by the device, the sample filtering task is triggered to filter the samples and augment the local dataset. Then, if the device status satisfies a pre-defined condition (\eg, the size of the augmented samples reaches a threshold), the training task is triggered to accomplish personalized model learning. Regarding the inference task, it is triggered by the app whenever the app needs to provide recommendations for the user.

\subsubsection{App}
In the mobile app, some pages contain dynamic contents for recommendation. When the user visits these certain pages, the event tracking system detects this event and sends a signal to trigger the model inference task. Based on the returned prediction, the app perceives the user's interest or intention and customizes the page accordingly to do real-time recommendation. Then, the following user behaviors in this time period are tracked and collected to form new device log. The log is aggregated and reported to the cloud and serves as the source of new samples in the next work cycle.

\section{Online Experiments in Taobao} \label{sec:experiments}
We have deployed the proposed \algname algorithm and the device-tunnel-cloud pipeline in Mobile Taobao app. We further extensively evaluate the online performance of our design and compare it with several baselines using A/B testing. We also report practical resource consumption to validate efficiency.

\subsection{Scenario, Task, and Model}\label{subsec:exp:scenario}

\begin{figure}[!t]
    \centering
    \includegraphics[width=0.8\linewidth]{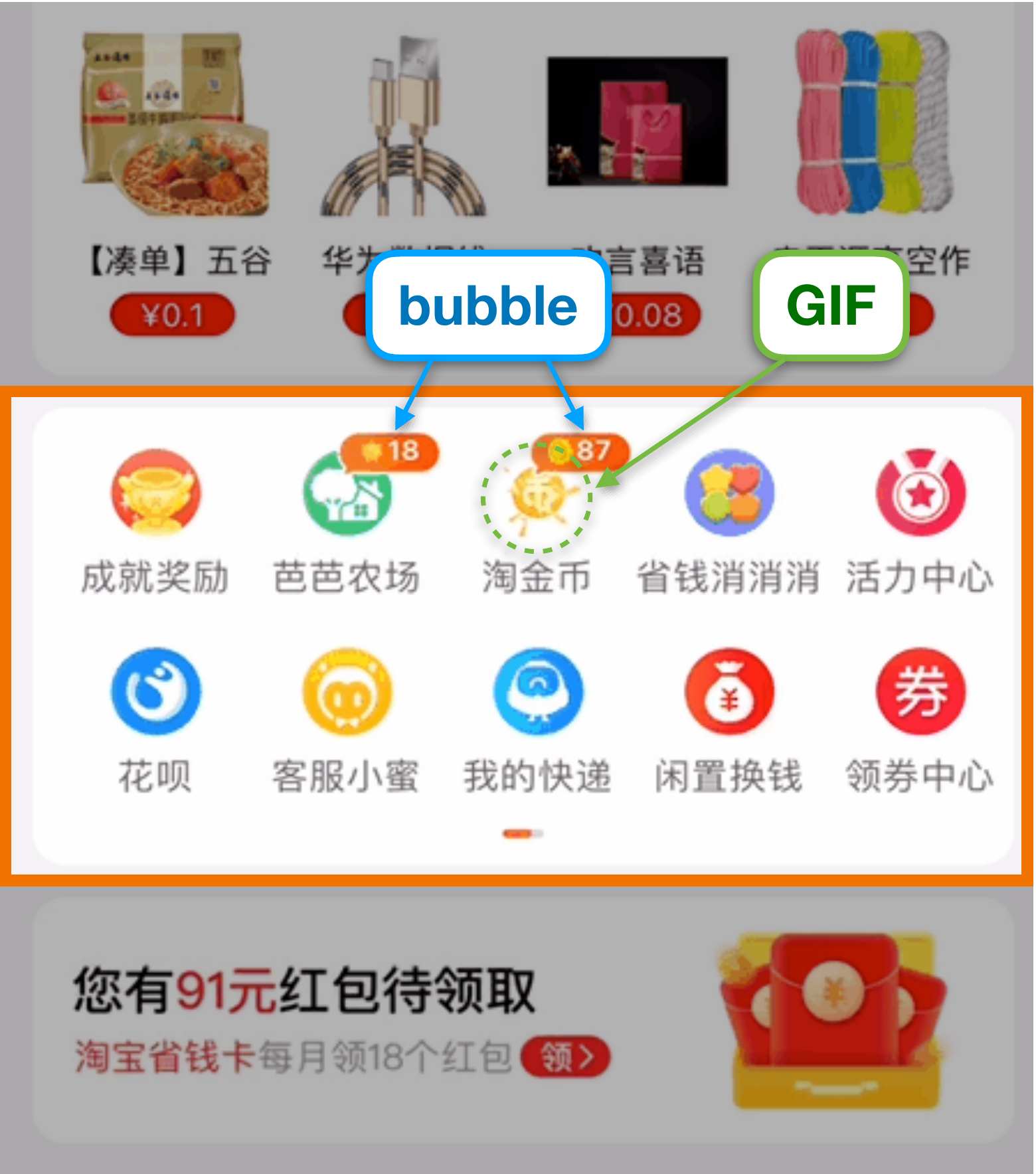}
    \caption{The icon area in Mobile Taobao.}
    \label{fig:taobao_icon_area}
\end{figure}

\begin{table}[!t]
    \centering
    \caption{Input and output of the icon CTR prediction task.}
    \label{tab:icon_inputs_outputs}
    \resizebox{\linewidth}{!}{
    \begin{tabular}[t]{clp{0.7\linewidth}}
        \toprule
        \multicolumn{1}{c}{\textbf{Type}} & \multicolumn{1}{c}{\textbf{Data Field}} & \multicolumn{1}{c}{\textbf{Meaning}}\\
        \midrule
         & target icon & The ID of the target icon.\\
        \cmidrule{2-3}
        & animation type & Bubble, GIF, or None.\\
        \cmidrule{2-3}
        \multirow{7}{*}{\textbf{Input}} & profile & The profile of the user.\\
        \cmidrule{2-3}
        ~ & \multirow{2}{*}{\tabincell{c}{icon click\\ sequence}} & The recent 7-day historical sequence of clicked icons.\\
        \cmidrule{2-3}
        ~ & biz statistics & The recent 7-day statistics on business visits.\\
        \cmidrule{2-3}
        ~ & \multirow{3}{*}{\tabincell{c}{behavior\\ sequence}} & The 512-length historical behavior sequence, including page names and behaviors (\eg, page views, button clicks, page leaves, etc.).\\
        \cmidrule{2-3}
        ~ & \multirow{2}{*}{\tabincell{c}{behavior\\ statistics}} & The 1-day statistics on user behaviors (\eg, click counts, add-to-cart counts, order counts, etc.).\\
        \midrule
        \textbf{Output} & CTR & The click through rate.\\
        \bottomrule
    \end{tabular}%

    }
\end{table}%

As shown in Figure~\ref{fig:taobao_icon_area}, the deployment scenario in Mobile Taobao is the icon area in the ``My Taobao'' page. This scenario contains 10 icons, and each icon is a shortcut to a certain business or function page. To attract user attention and increase participation, some animation effects (i.e., two bubbles and two GIFs) can be added over the icons. Further, the policy of choosing which icons to add animation effects depends mainly on the CTRs of the icons. Thus, the learning task is to accurately predict the CTR of each icon. 

We list the detailed input and output data fields of the CTR prediction task in Table~\ref{tab:icon_inputs_outputs}. In particular, the input data fields include the ID of the target icon, the type of animation, the user profile, the historical sequence of clicked icons, the user behavior sequence, as well as the statistics on the business visits and user behaviors.

\begin{figure}[!t]
    \centering
    \includegraphics[width=0.9\linewidth]{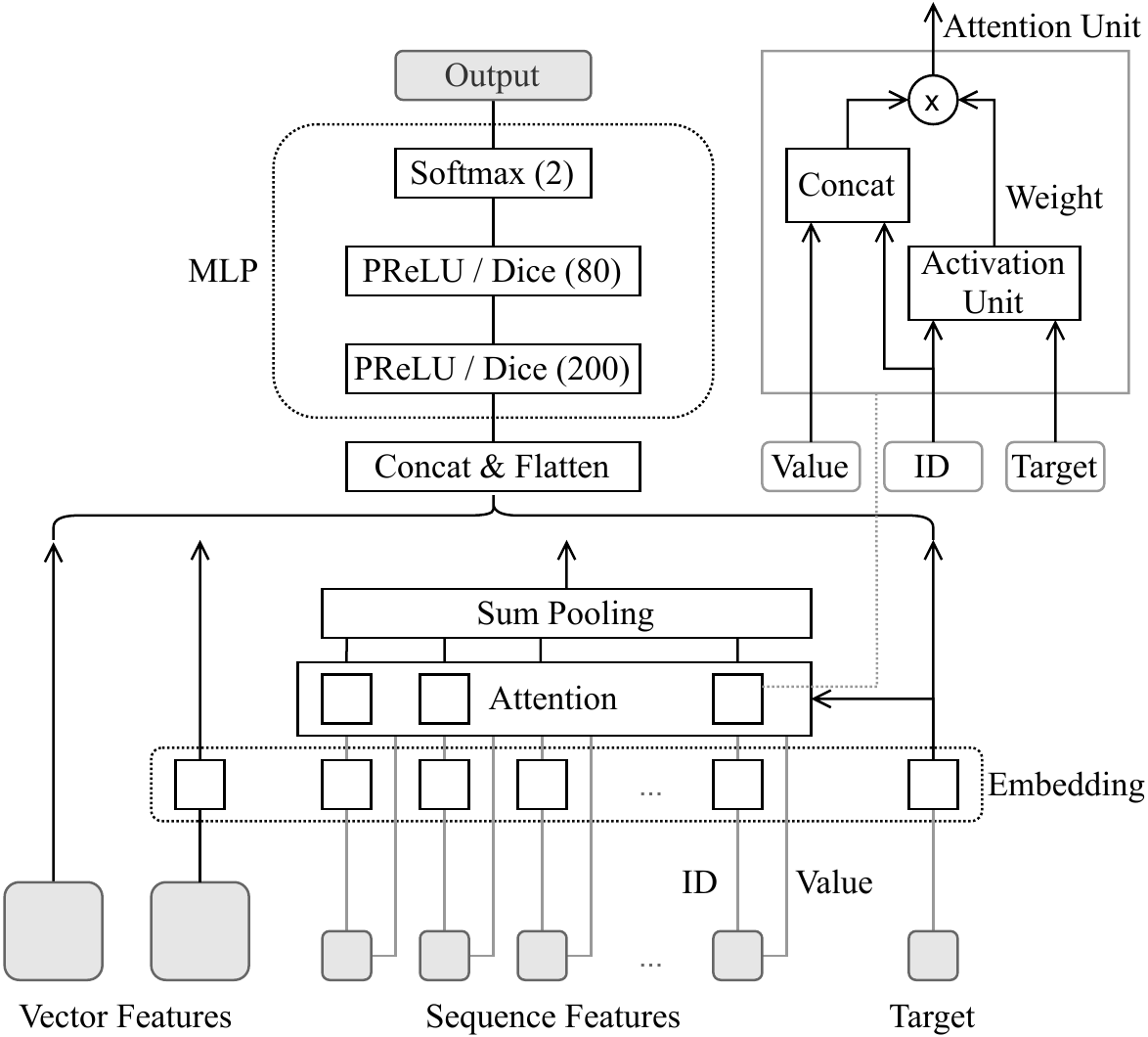}
    \caption{The DIN structure of the CTR prediction model.}
    \label{fig:model_structure}
\end{figure}

We take DIN as the CTR prediction model as shown in Figure~\ref{fig:model_structure}. It contains an embedding layer, where the embedding dimension is set to 3, an attention layer, and fully connected layers (\ie, MLP layers). We use the target icon to do attention with the sequence of clicked icons and the user behavior sequence. 

We still use DIN as the sample classifier for each user to distinguish the local and the outside samples. The key difference from the CTR prediction model is that only the user behavior sequence is used as the input, and thus, the embedding layer and the attention layer need to be pruned. Specifically, users' behavior sequences can reflect their behavior patterns, which are regarded as a good projection of the local data distribution. In contrast, the icon-related data fields can represent a user's preferences for different icons. If these data fields are also input to the sample classifier, the classifier will tend to search for the samples containing specific icons, which deviates from the original intention of finding outside samples that share a similar data distribution with the local samples, rather than the samples with the same targets.

\begin{figure*}[!t]
	\centering
	\includegraphics[width=.3\textwidth]{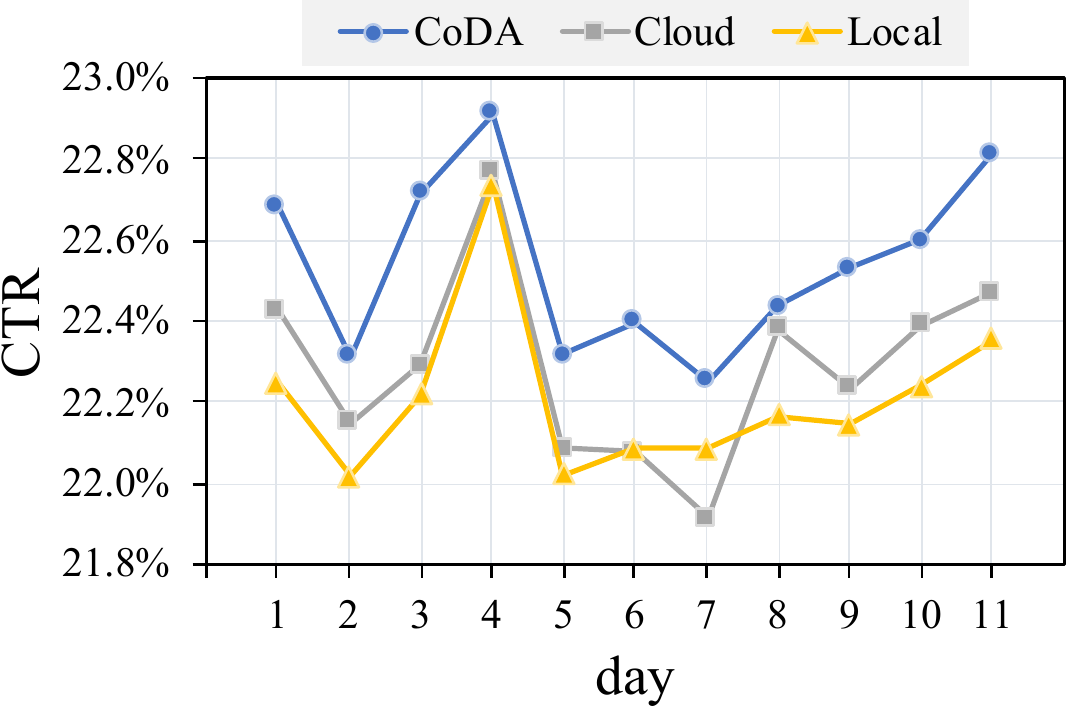}
	\hspace{0.02\linewidth}
    \includegraphics[width=.3\textwidth]{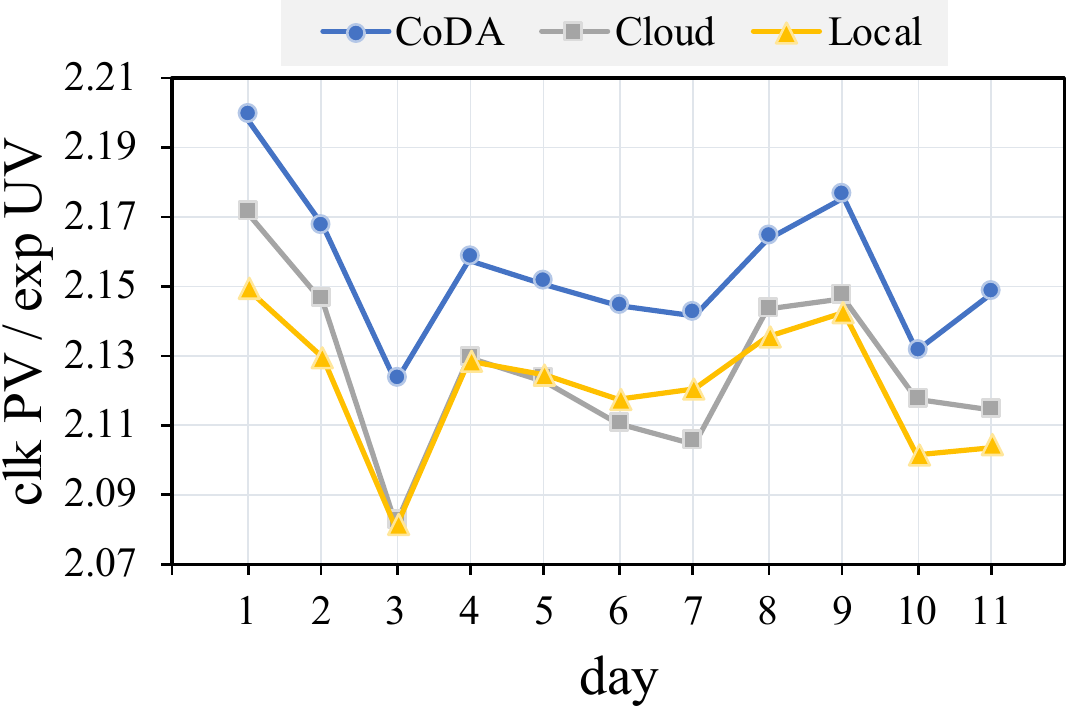}
	\hspace{0.02\linewidth}
    \includegraphics[width=.3\textwidth]{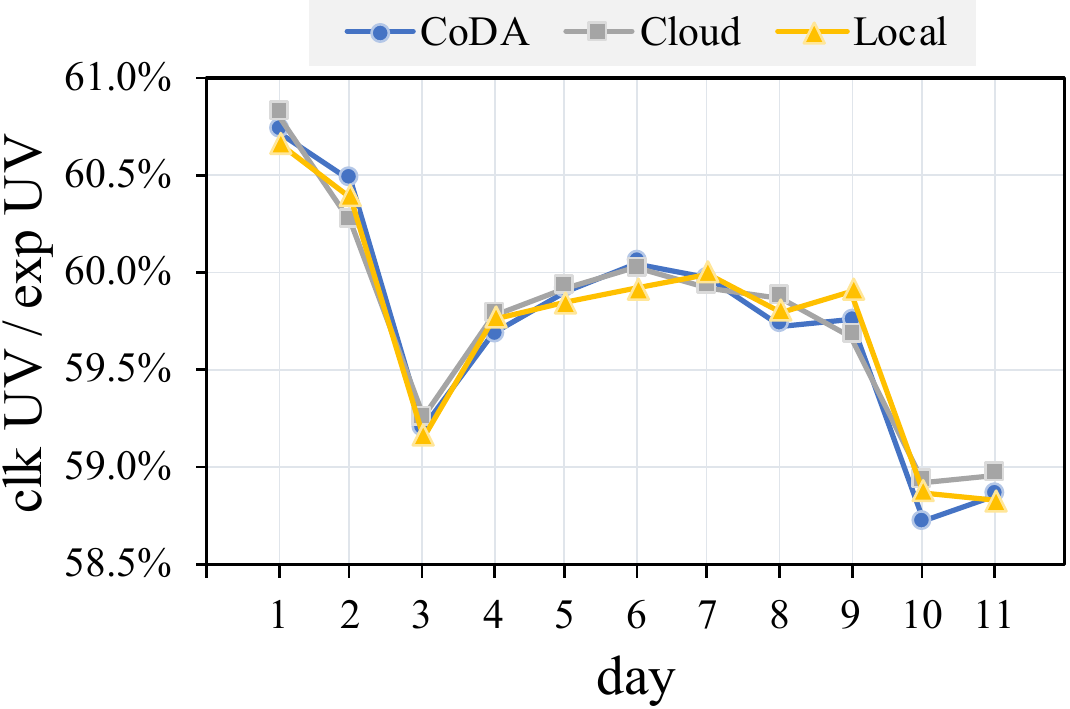}
	\caption{Online performance of our \algname and two baselines (i.e., cloud-based learning and local training) in Phase 1.}\label{fig:model_performance_20210722}
\end{figure*}

\begin{figure*}[!t]
	\centering
	\includegraphics[width=.3\textwidth]{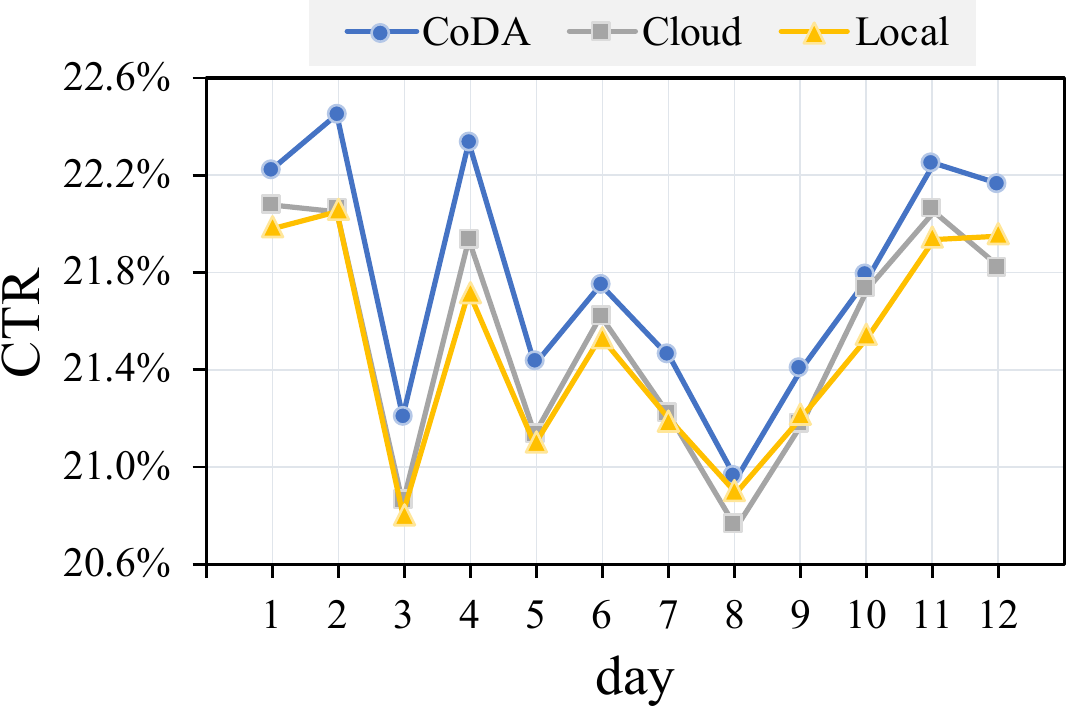}
	\hspace{0.02\linewidth}
    \includegraphics[width=.3\textwidth]{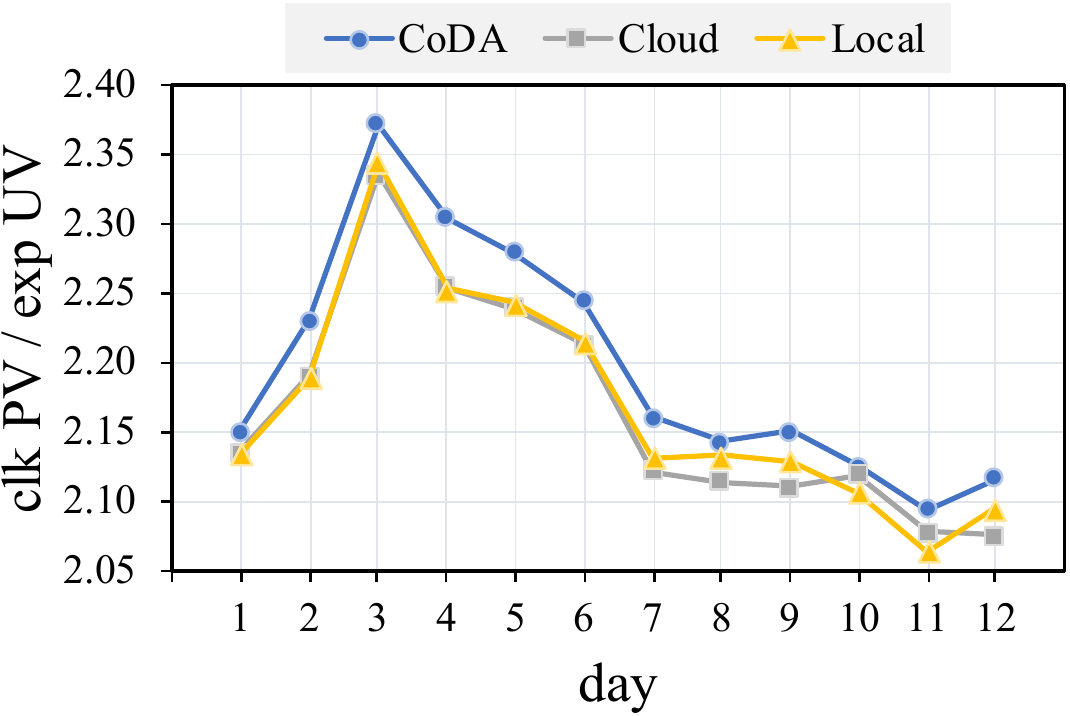}
	\hspace{0.02\linewidth}
    \includegraphics[width=.3\textwidth]{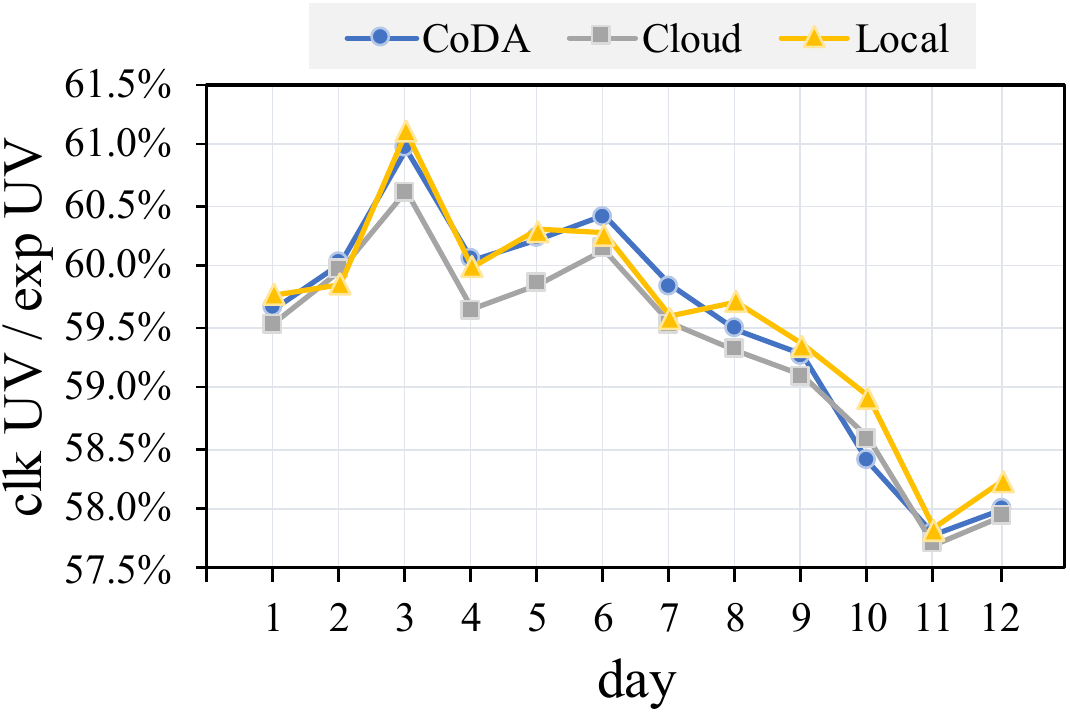}
    \caption{Online performance of our \algname and two baselines (i.e., cloud-based learning and local training) in Phase 2.}\label{fig:model_performance_20210804}
\end{figure*}

\subsection{Baselines, Metrics, and Setups}

We introduce two baselines for ablation study: (1) one is the initial global model, which is trained over the global data on the cloud. This baseline is currently the mainstream in industry and is introduced to verify the necessity of on-device training; and (2) the other is the on-device training over only the local data, which is to validate the necessity and the effectiveness of cloud-coordinated data augmentation.

We use three online metrics to rigorously evaluate the performance of different methods. Before giving the formal definitions, we first introduce some common abbreviations in recommender systems: ``clk'' is short for ``click''; ``exp'' is short for ``exposure''; ``PV'' is short for ``Page View''; ``UV'' is short for ``Unique Visitor''. Then, the three metrics are defined as follows: (1) $\frac{{\rm\bf clk\ PV}}{{\rm\bf exp\ PV}}$ is in fact {\bf CTR} and is the optimization objective of the learning task in our application scenario. CTR is also one of the most important metrics in recommender systems; (2) $\frac{{\rm\bf clk\ PV}}{{\rm\bf exp\ UV}}$ denotes the average number of clicks per user, which can measure the user activeness; and (3) $\frac{{\rm\bf clk\ UV}}{{\rm\bf exp\ UV}}$ denotes the percentage of the users who click the icon area at least once after exposure, which can capture the user liveness.

We finally introduce the deployment details: (1) For A/B testing, we create three non-overlapping buckets to deploy our \algname and two baselines. Each bucket roughly contains 100,000 randomly chosen highly active users with iOS devices. The model structure and input features keep the same for all three buckets; (2) on the side of cloud, the parameter $K$ in the KNN-based sample matching is set to 100, which indicates that 100 other users are matched for each user every day. The user-related feature used for sample matching is the biz statistics feature listed in Table~\ref{tab:icon_inputs_outputs}. In addition, to save cloud storage, only the matching results in the last 7 days are kept. Moreover, considering that the sample is in the format of sparse vectors, we compress the binary data with zlib to reduce size. The compressed binary data is further encoded into BASE64 strings to avoid being mishandled as control characters by network protocols and causing unexpected bugs. Such compression and encoding operations roughly reduce $90\%$ of data size, which not only dramatically saves the cloud storage but also sharply cuts down the communication overhead. The strings will be decompressed and recovered to samples automatically upon arriving at the smart devices; (3) regarding communication, to control workload and communication overhead, each device can pull at most 12 batches of matched samples every day. The number of samples in each batch is set to 25 by default, while on special occasions, at most 40 samples can be put in a batch to avoid batch fragments; and (4) on the side of device, regarding the CTR prediction model, we use the cloud-based training model as the initialization for it. After receiving the matched samples from the cloud, $1/3$ of them are used for the training of the sample classifier and the other $2/3$ are used for testing and further being filtered to do data augmentation. The filtering threshold $\sigma$ in the classifier-based sample filtering is set to 0.2. The sizes of the local samples and the filtered samples saved into the on-device database are each limited to be no more than 200. The threshold for triggering the training of the CTR prediction model is when the size of the augmented samples reaches 100. If a device has fetched all the matched samples available today (i.e., either no more samples are available on the cloud, or today's batch limit is reached), the training of the CTR prediction model will also be triggered to finish today's job. The parameters $t$ and $t'$ of the local sample lifecycle introduced in Section \ref{subsubsec:sys_resource_management} are set to 3 and 7, respectively. 

\subsection{Online Performance}

We conduct online A/B testing in two phases. Phase 1 lasts for 11 days in July 2021, and Phase 2 lasts for 12 days in August 2021. 

We plot the day-level results of Phase 1 and Phase 2 in Figure~\ref{fig:model_performance_20210722} and Figure~\ref{fig:model_performance_20210804}, respectively, which are consistent and stable in general. One key observation from Figure~\ref{fig:model_performance_20210722} and Figure~\ref{fig:model_performance_20210804} is that in terms of CTR and $\frac{{\rm clk\ PV}}{{\rm exp\ UV}}$, our \algname strictly outperforms both cloud-based learning and local training in all days, while the local training is slightly worse than the cloud-based learning. These results demonstrate that: (1) even though the local model is initialized with the global model trained over the global data on the cloud, on-device training over only the local data still falls into the classical dilemma of few-shot learning, causes severe overfitting problems, and degrades the model performance; and (2) by augmenting the local data with similar outside data using the sample classifier, \algname effectively mitigates overfitting and improves recommendation accuracy and user activeness. The second key observation is that in terms of $\frac{{\rm clk\ UV}}{{\rm exp\ UV}}$, the performance of three methods are quite close. This implies that on-device training and data augmentation may have little effect on the user liveness, particularly maintaining old click users and attracting new clock users.

\begin{table}[!t]
    \centering
    \caption{Accumulative online performance in Phase 1.}
    \label{tab:performance_data_20210722}
    \resizebox{\linewidth}{!}{
    \begin{tabular}[t]{cccccc}
        \toprule
             & \textbf{Cloud} & \textbf{Local} & \textbf{\algname} & \tabincell{c}{\textbf{\algname vs.} \\ \textbf{Cloud}} & \tabincell{c}{\textbf{\algname  vs.} \\ \textbf{Local}}\\
        \midrule
        \textbf{CTR} & 22.289\% & 22.209\% & 22.542\% & \textcolor{red}{\bf +1.13\%} & \textcolor{red}{\bf +1.50\%}\\
        \midrule
        $\frac{{\rm \bf clk\ PV}}{{\rm \bf exp\ UV}}$ & 18.978 & 18.919 & 19.267 & \textcolor{red}{\bf +1.52\%} & \textcolor{red}{\bf +1.84\%}\\ 
        \midrule
        $\frac{{\rm \bf clk\ UV}}{{\rm \bf exp\ UV}}$ & 85.150\% & 84.940\% & 85.141\% & -0.01\% & +0.24\%\\ 
        \bottomrule
    \end{tabular}
    }
\end{table}

\begin{table}[!t]
    \centering
    \caption{Accumulative online performance in Phase 2.}
    \label{tab:performance_data_20210804}
    \resizebox{\linewidth}{!}{
    \begin{tabular}[t]{cccccc}
        \toprule
            & \textbf{Cloud} & \textbf{Local} & \textbf{\algname} & \tabincell{c}{\textbf{\algname vs. }\\ \textbf{Cloud}} & \tabincell{c}{\textbf{\algname vs. }\\ \textbf{Local}}\\
        \midrule
        \textbf{CTR} & 21.517\% & 21.476\% & 21.765\% & \textcolor{red}{\bf +1.15\%} & \textcolor{red}{\bf +1.35\%}\\
        \midrule
        $\frac{{\rm \bf clk\ PV}}{{\rm \bf exp\ UV}}$ & 21.252 & 21.314 & 21.618 & \textcolor{red}{\bf +1.72\%} & \textcolor{red}{\bf +1.43\%}\\ 
        \midrule
        $\frac{{\rm \bf clk\ UV}}{{\rm \bf exp\ UV}}$ & 85.863\% & 85.992\% & 85.991\% & +0.15\% & -0.00\%\\ 
        \bottomrule
    \end{tabular}
    }
\end{table}

We also present the accumulated performance of Phase 1 and Phase 2 in Table~\ref{tab:performance_data_20210722} and Table~\ref{tab:performance_data_20210804}, respectively. Compared with the initial model using cloud-based training, \algname improves CTR and $\frac{{\rm clk\ PV}}{{\rm exp\ UV}}$ more than 1.1\% in both phases; compared with pure local training, the advantage of \algname is more remarkable, improving CTR and $\frac{{\rm clk\ PV}}{{\rm exp\ UV}}$ more than 1.3\% in both phases. Regarding $\frac{{\rm clk\ UV}}{{\rm exp\ UV}}$, the performances of three methods tie with each other.

The online A/B testing results above adequately demonstrate the necessity and effectiveness of integrating on-device training and data augmentation in the design of \algname.

\begin{figure}[!t]
    \centering
    \includegraphics[width=0.95\linewidth]{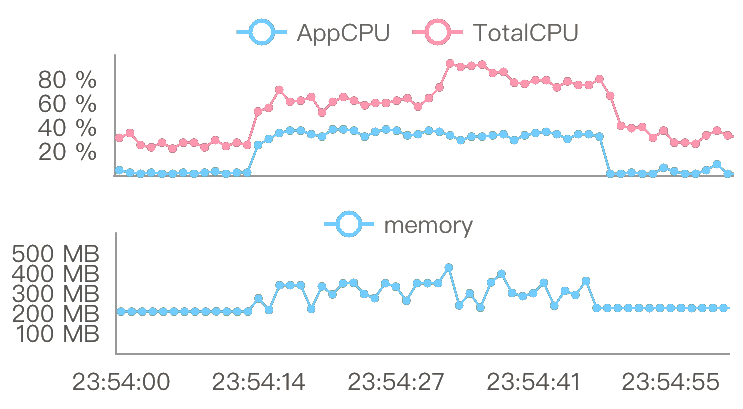}
    \caption{CPU and memory occupations of iPhone 8 Plus when executing \algname's on-device tasks for 50 times. AppCPU and TotalCPU denote the CPU usages of Mobile Taobao and the whole device system, respectively, while memory denotes the memory occupation of Mobile Taobao.}
    \label{fig:cpu_memory_consumption}
\end{figure}

\subsection{Resource Consumption}
In addition to the online performance of recommendation, the efficiency of executing device-related tasks is also a key consideration in practice. We use an iPhone 8 Plus as the test bed and install a debug version of Mobile Taobao on it. In particular, iPhone 8 Plus, released in 2017, is equipped with Apple A11 SoC (2 high-performance cores with 2.53 GHz maximum frequency and 4 high-efficiency cores with 1.42 GHz maximum frequency) and 3 GB LPDDR4X RAM, and falls into the midrange devices according to the Mobile Taobao's categorization in 2021. To evaluate the average resource consumption, we continuously trigger the device-related tasks for 50 times and repeat the test twice. We plot the CPU and memory occupations during one test in Figure~\ref{fig:cpu_memory_consumption}. We also report the major overhead in detail as follows: (1) one-round execution of \algname's on-device tasks costs 0.7 second in average. The task execution brings 35\% of CPU usage on one core at maximum; (2) roughly 120 MB additional memory is occupied during the task execution; and (3) the device downloads roughly 19.4 KB data for one trigger in average.

The light load of device-related tasks has validated the practical efficiency of our system design.

\section{Conclusion} \label{sec:conclustion}
In this work, we study model personalization in recommender systems. We have proposed an on-device personalized learning algorithm with cloud-coordinated data augmentation, called \algname, to breaks the dilemmas of purely cloud-based learning and on-device learning. We also have built a cloud-tunnel-device pipeline to support the workflow of \algname. We have deployed \algname in a practical recommendation scenario of Mobile Taobao. Online A/B testing and real device testing demonstrate the effectiveness and efficiency of \algname.

\bibliographystyle{ACM-Reference-Format}
\bibliography{reference}


\begin{thebibliography}{49}


\ifx \showCODEN    \undefined \def \showCODEN     #1{\unskip}     \fi
\ifx \showDOI      \undefined \def \showDOI       #1{#1}\fi
\ifx \showISBNx    \undefined \def \showISBNx     #1{\unskip}     \fi
\ifx \showISBNxiii \undefined \def \showISBNxiii  #1{\unskip}     \fi
\ifx \showISSN     \undefined \def \showISSN      #1{\unskip}     \fi
\ifx \showLCCN     \undefined \def \showLCCN      #1{\unskip}     \fi
\ifx \shownote     \undefined \def \shownote      #1{#1}          \fi
\ifx \showarticletitle \undefined \def \showarticletitle #1{#1}   \fi
\ifx \showURL      \undefined \def \showURL       {\relax}        \fi
\providecommand\bibfield[2]{#2}
\providecommand\bibinfo[2]{#2}
\providecommand\natexlab[1]{#1}
\providecommand\showeprint[2][]{arXiv:#2}

\bibitem[\protect\citeauthoryear{Abadi, Agarwal, Barham, et~al\mbox{.}}{Abadi
  et~al\mbox{.}}{2015}]%
        {tensorflow2015-whitepaper}
\bibfield{author}{\bibinfo{person}{Mart\'{\i}n Abadi}, \bibinfo{person}{Ashish
  Agarwal}, \bibinfo{person}{Paul Barham}, {et~al\mbox{.}}}
  \bibinfo{year}{2015}\natexlab{}.
\newblock \bibinfo{title}{{TensorFlow}: Large-Scale Machine Learning on
  Heterogeneous Systems}.
\newblock
\newblock
\urldef\tempurl%
\url{https://www.tensorflow.org/}
\showURL{%
\tempurl}
\newblock
\shownote{Software available from tensorflow.org.}


\bibitem[\protect\citeauthoryear{Amazon}{Amazon}{2021}]%
        {Amazon_2020_results}
\bibfield{author}{\bibinfo{person}{Amazon}.} \bibinfo{year}{2021}\natexlab{}.
\newblock \bibinfo{title}{Amazon.com Announces Financial Results and CEO
  Transition}.
\newblock
  \bibinfo{howpublished}{\url{https://ir.aboutamazon.com/news-release/news-release-details/2021/Amazon.com-Announces-Fourth-Quarter-Results/}}.
\newblock
\newblock
\shownote{Last accessed: Sept. 7, 2021.}


\bibitem[\protect\citeauthoryear{Arjovsky, Chintala, and Bottou}{Arjovsky
  et~al\mbox{.}}{2017}]%
        {proc:icml17:WGAN}
\bibfield{author}{\bibinfo{person}{Martin Arjovsky}, \bibinfo{person}{Soumith
  Chintala}, {and} \bibinfo{person}{L{\'e}on Bottou}.}
  \bibinfo{year}{2017}\natexlab{}.
\newblock \showarticletitle{{W}asserstein Generative Adversarial Networks}. In
  \bibinfo{booktitle}{\emph{Proc. of ICML}}. \bibinfo{publisher}{PMLR},
  \bibinfo{pages}{214--223}.
\newblock


\bibitem[\protect\citeauthoryear{Bistritz, Mann, and Bambos}{Bistritz
  et~al\mbox{.}}{2020}]%
        {NEURIPS2020_DistributedDistillation}
\bibfield{author}{\bibinfo{person}{Ilai Bistritz}, \bibinfo{person}{Ariana
  Mann}, {and} \bibinfo{person}{Nicholas Bambos}.}
  \bibinfo{year}{2020}\natexlab{}.
\newblock \showarticletitle{Distributed Distillation for On-Device Learning}.
  In \bibinfo{booktitle}{\emph{{Proc. of NeurIPS}}}.
  \bibinfo{pages}{22593--22604}.
\newblock


\bibitem[\protect\citeauthoryear{Bobadilla, Ortega, Hernando, and
  Guti{\'{e}}rrez}{Bobadilla et~al\mbox{.}}{2013}]%
        {10.1016/j.knosys.2013.03.012}
\bibfield{author}{\bibinfo{person}{Jes{\'{u}}s Bobadilla},
  \bibinfo{person}{Fernando Ortega}, \bibinfo{person}{Antonio Hernando}, {and}
  \bibinfo{person}{Abraham Guti{\'{e}}rrez}.} \bibinfo{year}{2013}\natexlab{}.
\newblock \showarticletitle{Recommender systems survey}.
\newblock \bibinfo{journal}{\emph{Knowl. Based Syst.}}  \bibinfo{volume}{46}
  (\bibinfo{year}{2013}), \bibinfo{pages}{109--132}.
\newblock


\bibitem[\protect\citeauthoryear{Chae, Kang, Kim, and Choi}{Chae
  et~al\mbox{.}}{2019}]%
        {proc:www19:RAGANBT}
\bibfield{author}{\bibinfo{person}{Dong{-}Kyu Chae}, \bibinfo{person}{Jin{-}Soo
  Kang}, \bibinfo{person}{Sang{-}Wook Kim}, {and} \bibinfo{person}{Jaeho
  Choi}.} \bibinfo{year}{2019}\natexlab{}.
\newblock \showarticletitle{Rating Augmentation with Generative Adversarial
  Networks towards Accurate Collaborative Filtering}. In
  \bibinfo{booktitle}{\emph{Proc. of {WWW}}}. \bibinfo{publisher}{{ACM}},
  \bibinfo{pages}{2616--2622}.
\newblock


\bibitem[\protect\citeauthoryear{Chae, Kim, Chau, and Kim}{Chae
  et~al\mbox{.}}{2020}]%
        {proc:sigir20:AR-CF:gan}
\bibfield{author}{\bibinfo{person}{Dong{-}Kyu Chae}, \bibinfo{person}{Jihoo
  Kim}, \bibinfo{person}{Duen~Horng Chau}, {and} \bibinfo{person}{Sang{-}Wook
  Kim}.} \bibinfo{year}{2020}\natexlab{}.
\newblock \showarticletitle{{AR-CF:} Augmenting Virtual Users and Items in
  Collaborative Filtering for Addressing Cold-Start Problems}. In
  \bibinfo{booktitle}{\emph{Proc. of {SIGIR}}}. \bibinfo{publisher}{{ACM}},
  \bibinfo{pages}{1251--1260}.
\newblock


\bibitem[\protect\citeauthoryear{Chen, Luo, Dong, Li, and He}{Chen
  et~al\mbox{.}}{2018}]%
        {chen2018federated}
\bibfield{author}{\bibinfo{person}{Fei Chen}, \bibinfo{person}{Mi Luo},
  \bibinfo{person}{Zhenhua Dong}, \bibinfo{person}{Zhenguo Li}, {and}
  \bibinfo{person}{Xiuqiang He}.} \bibinfo{year}{2018}\natexlab{}.
\newblock \bibinfo{title}{Federated meta-learning with fast convergence and
  efficient communication}.
\newblock \bibinfo{howpublished}{arXiv: 1802.07876}.
\newblock
\newblock
\shownote{\url{http://arxiv.org/abs/1802.07876}.}


\bibitem[\protect\citeauthoryear{Cheng, Koc, Harmsen, Shaked, Chandra, Aradhye,
  Anderson, Corrado, Chai, Ispir, Anil, Haque, Hong, Jain, Liu, and Shah}{Cheng
  et~al\mbox{.}}{2016}]%
        {10.1145/2988450.2988454}
\bibfield{author}{\bibinfo{person}{Heng{-}Tze Cheng}, \bibinfo{person}{Levent
  Koc}, \bibinfo{person}{Jeremiah Harmsen}, \bibinfo{person}{Tal Shaked},
  \bibinfo{person}{Tushar Chandra}, \bibinfo{person}{Hrishi Aradhye},
  \bibinfo{person}{Glen Anderson}, \bibinfo{person}{Greg Corrado},
  \bibinfo{person}{Wei Chai}, \bibinfo{person}{Mustafa Ispir},
  \bibinfo{person}{Rohan Anil}, \bibinfo{person}{Zakaria Haque},
  \bibinfo{person}{Lichan Hong}, \bibinfo{person}{Vihan Jain},
  \bibinfo{person}{Xiaobing Liu}, {and} \bibinfo{person}{Hemal Shah}.}
  \bibinfo{year}{2016}\natexlab{}.
\newblock \showarticletitle{Wide {\&} Deep Learning for Recommender Systems}.
  In \bibinfo{booktitle}{\emph{Proc. of the 1st Workshop on Deep Learning for
  Recommender Systems}}. \bibinfo{publisher}{{ACM}}, \bibinfo{pages}{7--10}.
\newblock


\bibitem[\protect\citeauthoryear{Cloud}{Cloud}{2017}]%
        {MaxCompute}
\bibfield{author}{\bibinfo{person}{Alibaba Cloud}.}
  \bibinfo{year}{2017}\natexlab{}.
\newblock \bibinfo{title}{MaxCompute}.
\newblock
  \bibinfo{howpublished}{\url{https://www.alibabacloud.com/product/maxcompute}}.
\newblock
\newblock
\shownote{Last accessed: Aug. 27, 2021.}


\bibitem[\protect\citeauthoryear{Cloud}{Cloud}{2020}]%
        {PAI}
\bibfield{author}{\bibinfo{person}{Alibaba Cloud}.}
  \bibinfo{year}{2020}\natexlab{}.
\newblock \bibinfo{title}{Platform of Artificial Intelligence}.
\newblock
  \bibinfo{howpublished}{\url{https://www.alibabacloud.com/product/machine-learning}}.
\newblock
\newblock
\shownote{Last accessed: Aug. 27, 2021.}


\bibitem[\protect\citeauthoryear{Cubuk, Zoph, Mane, Vasudevan, and Le}{Cubuk
  et~al\mbox{.}}{2018}]%
        {cubuk2018autoaugment}
\bibfield{author}{\bibinfo{person}{Ekin~D Cubuk}, \bibinfo{person}{Barret
  Zoph}, \bibinfo{person}{Dandelion Mane}, \bibinfo{person}{Vijay Vasudevan},
  {and} \bibinfo{person}{Quoc~V Le}.} \bibinfo{year}{2018}\natexlab{}.
\newblock \bibinfo{title}{Autoaugment: Learning augmentation policies from
  data}.
\newblock \bibinfo{howpublished}{arXiv: 1805.09501}.
\newblock
\newblock
\shownote{\url{http://arxiv.org/abs/1805.09501}.}


\bibitem[\protect\citeauthoryear{Deng, Kamani, and Mahdavi}{Deng
  et~al\mbox{.}}{2020}]%
        {deng2020adaptive}
\bibfield{author}{\bibinfo{person}{Yuyang Deng},
  \bibinfo{person}{Mohammad~Mahdi Kamani}, {and} \bibinfo{person}{Mehrdad
  Mahdavi}.} \bibinfo{year}{2020}\natexlab{}.
\newblock \bibinfo{title}{Adaptive personalized federated learning}.
\newblock \bibinfo{howpublished}{arXiv: 2003.13461}.
\newblock
\newblock
\shownote{\url{http://arxiv.org/abs/2003.13461}.}


\bibitem[\protect\citeauthoryear{Eichner, Koren, McMahan, Srebro, and
  Talwar}{Eichner et~al\mbox{.}}{2019}]%
        {pmlr-v97-eichner19a}
\bibfield{author}{\bibinfo{person}{Hubert Eichner}, \bibinfo{person}{Tomer
  Koren}, \bibinfo{person}{H.~Brendan McMahan}, \bibinfo{person}{Nathan
  Srebro}, {and} \bibinfo{person}{Kunal Talwar}.}
  \bibinfo{year}{2019}\natexlab{}.
\newblock \showarticletitle{Semi-Cyclic Stochastic Gradient Descent}. In
  \bibinfo{booktitle}{\emph{{Proc. of ICML}}}. \bibinfo{publisher}{PMLR},
  \bibinfo{pages}{1764--1773}.
\newblock


\bibitem[\protect\citeauthoryear{Eshratifar, Abrishami, and Pedram}{Eshratifar
  et~al\mbox{.}}{2021}]%
        {8871124}
\bibfield{author}{\bibinfo{person}{Amir~Erfan Eshratifar},
  \bibinfo{person}{Mohammad~Saeed Abrishami}, {and} \bibinfo{person}{Massoud
  Pedram}.} \bibinfo{year}{2021}\natexlab{}.
\newblock \showarticletitle{JointDNN: An Efficient Training and Inference
  Engine for Intelligent Mobile Cloud Computing Services}.
\newblock \bibinfo{journal}{\emph{{IEEE} Trans. Mob. Comput.}}
  \bibinfo{volume}{20}, \bibinfo{number}{2} (\bibinfo{year}{2021}),
  \bibinfo{pages}{565--576}.
\newblock


\bibitem[\protect\citeauthoryear{Fallah, Mokhtari, and Ozdaglar}{Fallah
  et~al\mbox{.}}{2020a}]%
        {fallah2020personalized}
\bibfield{author}{\bibinfo{person}{Alireza Fallah}, \bibinfo{person}{Aryan
  Mokhtari}, {and} \bibinfo{person}{Asuman Ozdaglar}.}
  \bibinfo{year}{2020}\natexlab{a}.
\newblock \bibinfo{title}{Personalized federated learning: A meta-learning
  approach}.
\newblock \bibinfo{howpublished}{arXiv: 2002.07948}.
\newblock
\newblock
\shownote{\url{http://arxiv.org/abs/2002.07948}.}


\bibitem[\protect\citeauthoryear{Fallah, Mokhtari, and Ozdaglar}{Fallah
  et~al\mbox{.}}{2020b}]%
        {NEURIPS2020_24389bfe}
\bibfield{author}{\bibinfo{person}{Alireza Fallah}, \bibinfo{person}{Aryan
  Mokhtari}, {and} \bibinfo{person}{Asuman~E. Ozdaglar}.}
  \bibinfo{year}{2020}\natexlab{b}.
\newblock \showarticletitle{Personalized Federated Learning with Theoretical
  Guarantees: {A} Model-Agnostic Meta-Learning Approach}. In
  \bibinfo{booktitle}{\emph{Proc. of NeurIPS}}. \bibinfo{pages}{3557--3568}.
\newblock


\bibitem[\protect\citeauthoryear{Gong, Jiang, Feng, Hu, Zhao, Liu, and Ou}{Gong
  et~al\mbox{.}}{2020}]%
        {10.1145/3340531.3412700}
\bibfield{author}{\bibinfo{person}{Yu Gong}, \bibinfo{person}{Ziwen Jiang},
  \bibinfo{person}{Yufei Feng}, \bibinfo{person}{Binbin Hu},
  \bibinfo{person}{Kaiqi Zhao}, \bibinfo{person}{Qingwen Liu}, {and}
  \bibinfo{person}{Wenwu Ou}.} \bibinfo{year}{2020}\natexlab{}.
\newblock \showarticletitle{EdgeRec: Recommender System on Edge in Mobile
  Taobao}. In \bibinfo{booktitle}{\emph{Proc. of {CIKM}}}.
  \bibinfo{publisher}{{ACM}}, \bibinfo{pages}{2477--2484}.
\newblock


\bibitem[\protect\citeauthoryear{Goodfellow, Pouget{-}Abadie, Mirza, Xu,
  Warde{-}Farley, Ozair, Courville, and Bengio}{Goodfellow
  et~al\mbox{.}}{2014}]%
        {GAN}
\bibfield{author}{\bibinfo{person}{Ian~J. Goodfellow}, \bibinfo{person}{Jean
  Pouget{-}Abadie}, \bibinfo{person}{Mehdi Mirza}, \bibinfo{person}{Bing Xu},
  \bibinfo{person}{David Warde{-}Farley}, \bibinfo{person}{Sherjil Ozair},
  \bibinfo{person}{Aaron~C. Courville}, {and} \bibinfo{person}{Yoshua Bengio}.}
  \bibinfo{year}{2014}\natexlab{}.
\newblock \showarticletitle{Generative Adversarial Nets}. In
  \bibinfo{booktitle}{\emph{{Proc. of NeurIPS}}}. \bibinfo{pages}{2672--2680}.
\newblock


\bibitem[\protect\citeauthoryear{Group}{Group}{2021}]%
        {Alibaba_2021_results}
\bibfield{author}{\bibinfo{person}{Alibaba Group}.}
  \bibinfo{year}{2021}\natexlab{}.
\newblock \bibinfo{title}{Alibaba Group 2021 Annual and Transition Report}.
\newblock
  \bibinfo{howpublished}{\url{https://otp.investis.com/clients/us/alibaba/SEC/sec-show.aspx?Type=html&FilingId=15112567&Cik=0001577552}}.
\newblock
\newblock
\shownote{Last accessed: Sept. 7, 2021.}


\bibitem[\protect\citeauthoryear{Gu, Niu, Wu, Chen, Hu, Lyu, and Wu}{Gu
  et~al\mbox{.}}{2021}]%
        {jour:csur21:survey:gu}
\bibfield{author}{\bibinfo{person}{Renjie Gu}, \bibinfo{person}{Chaoyue Niu},
  \bibinfo{person}{Fan Wu}, \bibinfo{person}{Guihai Chen},
  \bibinfo{person}{Chun Hu}, \bibinfo{person}{Chengfei Lyu}, {and}
  \bibinfo{person}{Zhihua Wu}.} \bibinfo{year}{2021}\natexlab{}.
\newblock \showarticletitle{From Server-Based to Client-Based Machine Learning:
  {A} Comprehensive Survey}.
\newblock \bibinfo{journal}{\emph{{ACM} Comput. Surv.}} \bibinfo{volume}{54},
  \bibinfo{number}{1} (\bibinfo{year}{2021}), \bibinfo{pages}{6:1--6:36}.
\newblock


\bibitem[\protect\citeauthoryear{Guo, Tang, Ye, Li, and He}{Guo
  et~al\mbox{.}}{2017}]%
        {10.5555/3172077.3172127}
\bibfield{author}{\bibinfo{person}{Huifeng Guo}, \bibinfo{person}{Ruiming
  Tang}, \bibinfo{person}{Yunming Ye}, \bibinfo{person}{Zhenguo Li}, {and}
  \bibinfo{person}{Xiuqiang He}.} \bibinfo{year}{2017}\natexlab{}.
\newblock \showarticletitle{DeepFM: {A} Factorization-Machine based Neural
  Network for {CTR} Prediction}. In \bibinfo{booktitle}{\emph{Proc. of
  {IJCAI}}}. \bibinfo{publisher}{ijcai.org}, \bibinfo{pages}{1725--1731}.
\newblock


\bibitem[\protect\citeauthoryear{Jiang, Wang, Chen, Wu, Wang, Zou, Yang, Cui,
  Cai, Yu, Lyu, and Wu}{Jiang et~al\mbox{.}}{2020}]%
        {alibaba2020mnn}
\bibfield{author}{\bibinfo{person}{Xiaotang Jiang}, \bibinfo{person}{Huan
  Wang}, \bibinfo{person}{Yiliu Chen}, \bibinfo{person}{Ziqi Wu},
  \bibinfo{person}{Lichuan Wang}, \bibinfo{person}{Bin Zou},
  \bibinfo{person}{Yafeng Yang}, \bibinfo{person}{Zongyang Cui},
  \bibinfo{person}{Yu Cai}, \bibinfo{person}{Tianhang Yu},
  \bibinfo{person}{Chengfei Lyu}, {and} \bibinfo{person}{Zhihua Wu}.}
  \bibinfo{year}{2020}\natexlab{}.
\newblock \showarticletitle{{MNN:} {A} Universal and Efficient Inference
  Engine}. In \bibinfo{booktitle}{\emph{{Proc. of MLSys}}},
  Vol.~\bibinfo{volume}{2}. \bibinfo{pages}{1--13}.
\newblock


\bibitem[\protect\citeauthoryear{Jiang, Chang, and Wang}{Jiang
  et~al\mbox{.}}{2021}]%
        {tc:arxiv21:TransGAN}
\bibfield{author}{\bibinfo{person}{Yifan Jiang}, \bibinfo{person}{Shiyu Chang},
  {and} \bibinfo{person}{Zhangyang Wang}.} \bibinfo{year}{2021}\natexlab{}.
\newblock \bibinfo{title}{Transgan: Two transformers can make one strong gan}.
\newblock \bibinfo{howpublished}{arXiv: 2102.07074}.
\newblock
\newblock
\shownote{\url{http://arxiv.org/abs/2102.07074}.}


\bibitem[\protect\citeauthoryear{Jiang, Kone{\v{c}}n{\`y}, Rush, and
  Kannan}{Jiang et~al\mbox{.}}{2019}]%
        {jiang2019improving}
\bibfield{author}{\bibinfo{person}{Yihan Jiang}, \bibinfo{person}{Jakub
  Kone{\v{c}}n{\`y}}, \bibinfo{person}{Keith Rush}, {and}
  \bibinfo{person}{Sreeram Kannan}.} \bibinfo{year}{2019}\natexlab{}.
\newblock \bibinfo{title}{Improving federated learning personalization via
  model agnostic meta learning}.
\newblock \bibinfo{howpublished}{arXiv: 1909.12488}.
\newblock
\newblock
\shownote{\url{http://arxiv.org/abs/1909.12488}.}


\bibitem[\protect\citeauthoryear{Johnson, Douze, and J{\'e}gou}{Johnson
  et~al\mbox{.}}{2017}]%
        {Faiss}
\bibfield{author}{\bibinfo{person}{Jeff Johnson}, \bibinfo{person}{Matthijs
  Douze}, {and} \bibinfo{person}{Herv{\'e} J{\'e}gou}.}
  \bibinfo{year}{2017}\natexlab{}.
\newblock \bibinfo{title}{Billion-scale similarity search with GPUs}.
\newblock \bibinfo{howpublished}{arXiv: 1702.08734}.
\newblock
\newblock
\shownote{\url{http://arxiv.org/abs/1702.08734}.}


\bibitem[\protect\citeauthoryear{Kang, Dong, Zheng, and Yang}{Kang
  et~al\mbox{.}}{2017}]%
        {kang2017patchshuffle}
\bibfield{author}{\bibinfo{person}{Guoliang Kang}, \bibinfo{person}{Xuanyi
  Dong}, \bibinfo{person}{Liang Zheng}, {and} \bibinfo{person}{Yi Yang}.}
  \bibinfo{year}{2017}\natexlab{}.
\newblock \bibinfo{title}{Patchshuffle regularization}.
\newblock \bibinfo{howpublished}{arXiv: 1707.07103}.
\newblock
\newblock
\shownote{\url{http://arxiv.org/abs/1707.07103}.}


\bibitem[\protect\citeauthoryear{Karimireddy, Kale, Mohri, Reddi, Stich, and
  Suresh}{Karimireddy et~al\mbox{.}}{2020}]%
        {proc:icml20:scaffold}
\bibfield{author}{\bibinfo{person}{Sai~Praneeth Karimireddy},
  \bibinfo{person}{Satyen Kale}, \bibinfo{person}{Mehryar Mohri},
  \bibinfo{person}{Sashank~J. Reddi}, \bibinfo{person}{Sebastian~U. Stich},
  {and} \bibinfo{person}{Ananda~Theertha Suresh}.}
  \bibinfo{year}{2020}\natexlab{}.
\newblock \showarticletitle{{SCAFFOLD:} Stochastic Controlled Averaging for
  Federated Learning}. In \bibinfo{booktitle}{\emph{{Proc. of ICML}}},
  Vol.~\bibinfo{volume}{119}. \bibinfo{publisher}{{PMLR}},
  \bibinfo{pages}{5132--5143}.
\newblock


\bibitem[\protect\citeauthoryear{Li, Huang, Yang, Wang, and Zhang}{Li
  et~al\mbox{.}}{2019}]%
        {proc:iclr19:zhihuazhang:convergence}
\bibfield{author}{\bibinfo{person}{Xiang Li}, \bibinfo{person}{Kaixuan Huang},
  \bibinfo{person}{Wenhao Yang}, \bibinfo{person}{Shusen Wang}, {and}
  \bibinfo{person}{Zhihua Zhang}.} \bibinfo{year}{2019}\natexlab{}.
\newblock \showarticletitle{On the Convergence of {FedAvg} on Non-IID Data}. In
  \bibinfo{booktitle}{\emph{{Proc. of ICLR}}}.
  \bibinfo{publisher}{OpenReview.net}.
\newblock


\bibitem[\protect\citeauthoryear{Lin, Wang, Guan, Zhao, Zhao, Zhou, Li, and
  Qi}{Lin et~al\mbox{.}}{2019}]%
        {proc:pakdd20:RNE}
\bibfield{author}{\bibinfo{person}{Jianbin Lin}, \bibinfo{person}{Daixin Wang},
  \bibinfo{person}{Lu Guan}, \bibinfo{person}{Yin Zhao},
  \bibinfo{person}{Binqiang Zhao}, \bibinfo{person}{Jun Zhou},
  \bibinfo{person}{Xiaolong Li}, {and} \bibinfo{person}{Yuan~(Alan) Qi}.}
  \bibinfo{year}{2019}\natexlab{}.
\newblock \showarticletitle{{RNE:} {A} Scalable Network Embedding for
  Billion-Scale Recommendation}. In \bibinfo{booktitle}{\emph{Proc. of
  {PAKDD}}}. \bibinfo{publisher}{Springer}, \bibinfo{pages}{432--445}.
\newblock


\bibitem[\protect\citeauthoryear{Lin, Yang, and Zhang}{Lin
  et~al\mbox{.}}{2020}]%
        {9355664}
\bibfield{author}{\bibinfo{person}{Sen Lin}, \bibinfo{person}{Guang Yang},
  {and} \bibinfo{person}{Junshan Zhang}.} \bibinfo{year}{2020}\natexlab{}.
\newblock \showarticletitle{A Collaborative Learning Framework via Federated
  Meta-Learning}. In \bibinfo{booktitle}{\emph{Proc. of {ICDCS}}}.
  \bibinfo{publisher}{{IEEE}}, \bibinfo{pages}{289--299}.
\newblock


\bibitem[\protect\citeauthoryear{Lu, Shu, Tan, Liu, Zhou, Chen, and Pei}{Lu
  et~al\mbox{.}}{2019}]%
        {10.1145/3318216.3363304}
\bibfield{author}{\bibinfo{person}{Yan Lu}, \bibinfo{person}{Yuanchao Shu},
  \bibinfo{person}{Xu Tan}, \bibinfo{person}{Yunxin Liu},
  \bibinfo{person}{Mengyu Zhou}, \bibinfo{person}{Qi Chen}, {and}
  \bibinfo{person}{Dan Pei}.} \bibinfo{year}{2019}\natexlab{}.
\newblock \showarticletitle{Collaborative learning between cloud and end
  devices: an empirical study on location prediction}. In
  \bibinfo{booktitle}{\emph{Proc. of {SEC}}}. \bibinfo{publisher}{{ACM}},
  \bibinfo{pages}{139--151}.
\newblock


\bibitem[\protect\citeauthoryear{Mansour, Mohri, Ro, and Suresh}{Mansour
  et~al\mbox{.}}{2020}]%
        {mansour2020three}
\bibfield{author}{\bibinfo{person}{Yishay Mansour}, \bibinfo{person}{Mehryar
  Mohri}, \bibinfo{person}{Jae Ro}, {and} \bibinfo{person}{Ananda~Theertha
  Suresh}.} \bibinfo{year}{2020}\natexlab{}.
\newblock \bibinfo{title}{Three approaches for personalization with
  applications to federated learning}.
\newblock \bibinfo{howpublished}{arXiv: 2002.10619}.
\newblock
\newblock
\shownote{\url{http://arxiv.org/abs/2002.10619}.}


\bibitem[\protect\citeauthoryear{McMahan, Moore, Ramage, Hampson, and
  y~Arcas}{McMahan et~al\mbox{.}}{2017}]%
        {mcmahan2016communication}
\bibfield{author}{\bibinfo{person}{H.~Brendan McMahan}, \bibinfo{person}{Eider
  Moore}, \bibinfo{person}{Daniel Ramage}, \bibinfo{person}{Seth Hampson},
  {and} \bibinfo{person}{Blaise~Ag{\"{u}}era y Arcas}.}
  \bibinfo{year}{2017}\natexlab{}.
\newblock \showarticletitle{Communication-Efficient Learning of Deep Networks
  from Decentralized Data}. In \bibinfo{booktitle}{\emph{Proc. of {AISTATS}}}.
  \bibinfo{publisher}{PMLR}, \bibinfo{pages}{1273--1282}.
\newblock


\bibitem[\protect\citeauthoryear{Mirza and Osindero}{Mirza and
  Osindero}{2014}]%
        {tc:arxiv14:CGAN}
\bibfield{author}{\bibinfo{person}{Mehdi Mirza} {and} \bibinfo{person}{Simon
  Osindero}.} \bibinfo{year}{2014}\natexlab{}.
\newblock \bibinfo{title}{Conditional Generative Adversarial Nets}.
\newblock \bibinfo{howpublished}{arXiv: 1411.1784}.
\newblock
\newblock
\shownote{\url{http://arxiv.org/abs/1411.1784}.}


\bibitem[\protect\citeauthoryear{Mohri, Sivek, and Suresh}{Mohri
  et~al\mbox{.}}{2019}]%
        {pmlr-v97-mohri19a}
\bibfield{author}{\bibinfo{person}{Mehryar Mohri}, \bibinfo{person}{Gary
  Sivek}, {and} \bibinfo{person}{Ananda~Theertha Suresh}.}
  \bibinfo{year}{2019}\natexlab{}.
\newblock \showarticletitle{Agnostic Federated Learning}. In
  \bibinfo{booktitle}{\emph{{Proc. of ICML}}}. \bibinfo{publisher}{PMLR},
  \bibinfo{pages}{4615--4625}.
\newblock


\bibitem[\protect\citeauthoryear{Moreno{-}Barea, Strazzera, Jerez, Urda, and
  Franco}{Moreno{-}Barea et~al\mbox{.}}{2018}]%
        {moreno2018forward}
\bibfield{author}{\bibinfo{person}{Francisco~J. Moreno{-}Barea},
  \bibinfo{person}{Fiammetta Strazzera}, \bibinfo{person}{Jos{\'{e}}~M. Jerez},
  \bibinfo{person}{Daniel Urda}, {and} \bibinfo{person}{Leonardo Franco}.}
  \bibinfo{year}{2018}\natexlab{}.
\newblock \showarticletitle{Forward Noise Adjustment Scheme for Data
  Augmentation}. In \bibinfo{booktitle}{\emph{Proc. of {SSCI}}}.
  \bibinfo{publisher}{{IEEE}}, \bibinfo{pages}{728--734}.
\newblock


\bibitem[\protect\citeauthoryear{Radford, Metz, and Chintala}{Radford
  et~al\mbox{.}}{2016}]%
        {proc:iclr16:DCGAN}
\bibfield{author}{\bibinfo{person}{Alec Radford}, \bibinfo{person}{Luke Metz},
  {and} \bibinfo{person}{Soumith Chintala}.} \bibinfo{year}{2016}\natexlab{}.
\newblock \showarticletitle{Unsupervised Representation Learning with Deep
  Convolutional Generative Adversarial Networks}. In
  \bibinfo{booktitle}{\emph{Proc. of {ICLR}}}.
\newblock


\bibitem[\protect\citeauthoryear{Smith, Chiang, Sanjabi, and Talwalkar}{Smith
  et~al\mbox{.}}{2017}]%
        {proc:nips17:smith:fed:multitask}
\bibfield{author}{\bibinfo{person}{Virginia Smith}, \bibinfo{person}{Chao-Kai
  Chiang}, \bibinfo{person}{Maziar Sanjabi}, {and} \bibinfo{person}{Ameet
  Talwalkar}.} \bibinfo{year}{2017}\natexlab{}.
\newblock \showarticletitle{Federated Multi-Task Learning}. In
  \bibinfo{booktitle}{\emph{{Proc. of NeurIPS}}}. \bibinfo{pages}{4424--4434}.
\newblock


\bibitem[\protect\citeauthoryear{Wang, Huang, Zhao, Zhang, Zhao, and Lee}{Wang
  et~al\mbox{.}}{2018}]%
        {proc:kdd18:EGES}
\bibfield{author}{\bibinfo{person}{Jizhe Wang}, \bibinfo{person}{Pipei Huang},
  \bibinfo{person}{Huan Zhao}, \bibinfo{person}{Zhibo Zhang},
  \bibinfo{person}{Binqiang Zhao}, {and} \bibinfo{person}{Dik~Lun Lee}.}
  \bibinfo{year}{2018}\natexlab{}.
\newblock \showarticletitle{Billion-scale Commodity Embedding for E-commerce
  Recommendation in Alibaba}. In \bibinfo{booktitle}{\emph{Proc. of {KDD}}}.
  \bibinfo{publisher}{{ACM}}, \bibinfo{pages}{839--848}.
\newblock


\bibitem[\protect\citeauthoryear{Wang, Yin, Wang, Nguyen, Huang, and Cui}{Wang
  et~al\mbox{.}}{2019}]%
        {10.1145/3292500.3330873}
\bibfield{author}{\bibinfo{person}{Qinyong Wang}, \bibinfo{person}{Hongzhi
  Yin}, \bibinfo{person}{Hao Wang}, \bibinfo{person}{Quoc Viet~Hung Nguyen},
  \bibinfo{person}{Zi Huang}, {and} \bibinfo{person}{Lizhen Cui}.}
  \bibinfo{year}{2019}\natexlab{}.
\newblock \showarticletitle{Enhancing Collaborative Filtering with Generative
  Augmentation}. In \bibinfo{booktitle}{\emph{Proc. of {KDD}}}.
  \bibinfo{publisher}{{ACM}}, \bibinfo{pages}{548--556}.
\newblock


\bibitem[\protect\citeauthoryear{Wang, Fu, Fu, and Wang}{Wang
  et~al\mbox{.}}{2017}]%
        {10.1145/3124749.3124754}
\bibfield{author}{\bibinfo{person}{Ruoxi Wang}, \bibinfo{person}{Bin Fu},
  \bibinfo{person}{Gang Fu}, {and} \bibinfo{person}{Mingliang Wang}.}
  \bibinfo{year}{2017}\natexlab{}.
\newblock \showarticletitle{Deep {\&} Cross Network for Ad Click Predictions}.
  In \bibinfo{booktitle}{\emph{Proc. of ADKDD}}. \bibinfo{publisher}{{ACM}},
  \bibinfo{pages}{12:1--12:7}.
\newblock


\bibitem[\protect\citeauthoryear{Yao, Wang, Jia, Han, Zhou, and Yang}{Yao
  et~al\mbox{.}}{2021}]%
        {10.1145/3447548.3467097}
\bibfield{author}{\bibinfo{person}{Jiangchao Yao}, \bibinfo{person}{Feng Wang},
  \bibinfo{person}{Kunyang Jia}, \bibinfo{person}{Bo Han},
  \bibinfo{person}{Jingren Zhou}, {and} \bibinfo{person}{Hongxia Yang}.}
  \bibinfo{year}{2021}\natexlab{}.
\newblock \showarticletitle{Device-Cloud Collaborative Learning for
  Recommendation}. In \bibinfo{booktitle}{\emph{Proc. of {KDD}}}.
  \bibinfo{publisher}{{ACM}}, \bibinfo{pages}{3865--3874}.
\newblock


\bibitem[\protect\citeauthoryear{Yu, Jin, and Yang}{Yu et~al\mbox{.}}{2019}]%
        {pmlr-v97-yu19d}
\bibfield{author}{\bibinfo{person}{Hao Yu}, \bibinfo{person}{Rong Jin}, {and}
  \bibinfo{person}{Sen Yang}.} \bibinfo{year}{2019}\natexlab{}.
\newblock \showarticletitle{On the Linear Speedup Analysis of Communication
  Efficient Momentum {SGD} for Distributed Non-Convex Optimization}. In
  \bibinfo{booktitle}{\emph{{Proc. of ICML}}}. \bibinfo{publisher}{PMLR},
  \bibinfo{pages}{7184--7193}.
\newblock


\bibitem[\protect\citeauthoryear{Yuan, He, Karatzoglou, and Zhang}{Yuan
  et~al\mbox{.}}{2020}]%
        {proc:sigir20:patch}
\bibfield{author}{\bibinfo{person}{Fajie Yuan}, \bibinfo{person}{Xiangnan He},
  \bibinfo{person}{Alexandros Karatzoglou}, {and} \bibinfo{person}{Liguang
  Zhang}.} \bibinfo{year}{2020}\natexlab{}.
\newblock \showarticletitle{Parameter-Efficient Transfer from Sequential
  Behaviors for User Modeling and Recommendation}. In
  \bibinfo{booktitle}{\emph{{SIGIR}}}. \bibinfo{publisher}{{ACM}},
  \bibinfo{pages}{1469--1478}.
\newblock


\bibitem[\protect\citeauthoryear{Zhao, Li, Lai, Suda, Civin, and Chandra}{Zhao
  et~al\mbox{.}}{2018}]%
        {zhao2018federated}
\bibfield{author}{\bibinfo{person}{Yue Zhao}, \bibinfo{person}{Meng Li},
  \bibinfo{person}{Liangzhen Lai}, \bibinfo{person}{Naveen Suda},
  \bibinfo{person}{Damon Civin}, {and} \bibinfo{person}{Vikas Chandra}.}
  \bibinfo{year}{2018}\natexlab{}.
\newblock \bibinfo{title}{Federated learning with non-iid data}.
\newblock \bibinfo{howpublished}{arXiv: 1806.00582}.
\newblock
\newblock
\shownote{\url{http://arxiv.org/abs/1806.00582}.}


\bibitem[\protect\citeauthoryear{Zhong, Zheng, Kang, Li, and Yang}{Zhong
  et~al\mbox{.}}{2020}]%
        {zhong2020random}
\bibfield{author}{\bibinfo{person}{Zhun Zhong}, \bibinfo{person}{Liang Zheng},
  \bibinfo{person}{Guoliang Kang}, \bibinfo{person}{Shaozi Li}, {and}
  \bibinfo{person}{Yi Yang}.} \bibinfo{year}{2020}\natexlab{}.
\newblock \showarticletitle{Random Erasing Data Augmentation}. In
  \bibinfo{booktitle}{\emph{Proc. of {AAAI}}}. \bibinfo{publisher}{{AAAI}
  Press}, \bibinfo{pages}{13001--13008}.
\newblock


\bibitem[\protect\citeauthoryear{Zhou, Mou, Fan, Pi, Bian, Zhou, Zhu, and
  Gai}{Zhou et~al\mbox{.}}{2019}]%
        {proc:aaai19:DIEN}
\bibfield{author}{\bibinfo{person}{Guorui Zhou}, \bibinfo{person}{Na Mou},
  \bibinfo{person}{Ying Fan}, \bibinfo{person}{Qi Pi}, \bibinfo{person}{Weijie
  Bian}, \bibinfo{person}{Chang Zhou}, \bibinfo{person}{Xiaoqiang Zhu}, {and}
  \bibinfo{person}{Kun Gai}.} \bibinfo{year}{2019}\natexlab{}.
\newblock \showarticletitle{Deep Interest Evolution Network for Click-Through
  Rate Prediction}. In \bibinfo{booktitle}{\emph{Proc. of {AAAI}}}.
  \bibinfo{publisher}{{AAAI} Press}, \bibinfo{pages}{5941--5948}.
\newblock


\bibitem[\protect\citeauthoryear{Zhou, Zhu, Song, Fan, Zhu, Ma, Yan, Jin, Li,
  and Gai}{Zhou et~al\mbox{.}}{2018}]%
        {DIN}
\bibfield{author}{\bibinfo{person}{Guorui Zhou}, \bibinfo{person}{Xiaoqiang
  Zhu}, \bibinfo{person}{Chengru Song}, \bibinfo{person}{Ying Fan},
  \bibinfo{person}{Han Zhu}, \bibinfo{person}{Xiao Ma},
  \bibinfo{person}{Yanghui Yan}, \bibinfo{person}{Junqi Jin},
  \bibinfo{person}{Han Li}, {and} \bibinfo{person}{Kun Gai}.}
  \bibinfo{year}{2018}\natexlab{}.
\newblock \showarticletitle{Deep Interest Network for Click-Through Rate
  Prediction}. In \bibinfo{booktitle}{\emph{Proc. of {KDD}}}.
  \bibinfo{publisher}{ACM}, \bibinfo{pages}{1059--1068}.
\newblock


\end{thebibliography}

\end{document}